\documentclass[times,twocolumn,final]{elsarticle}

\usepackage{medima}
\usepackage{framed,multirow}

\usepackage{amssymb}
\usepackage{latexsym}

\usepackage{url}
\usepackage{xcolor}

\usepackage{hyperref}
\usepackage{amsmath,amssymb,amsfonts}
\usepackage{algorithmic}
\usepackage{graphicx}
\usepackage{textcomp}

\usepackage{chngcntr}
\usepackage{placeins}
\usepackage{amsfonts}       
\usepackage{nicefrac} 
\usepackage{amsmath,amssymb} 
\usepackage{amstext}
\usepackage{multirow}
\usepackage{tabu}
\usepackage{svg}
\usepackage{bbm}
\usepackage{diagbox}
\usepackage[english]{babel}
\usepackage{comment}
\usepackage{array}
\usepackage{amstext}
\usepackage{makecell}
\usepackage{caption}
\usepackage{tablefootnote}
\usepackage{babel}
\usepackage{booktabs} 
\usepackage{float}
\usepackage{graphicx}
\usepackage{multirow}
\usepackage{booktabs}
\usepackage{multirow}
\usepackage{siunitx}
\usepackage[table]{xcolor}
\usepackage{adjustbox}
\usepackage{enumitem}
\usepackage{booktabs}
\usepackage{arydshln}
\usepackage{stfloats}
\definecolor{lightgray}{gray}{0.92}
\usepackage{subcaption} 

\definecolor{newcolor}{rgb}{.8,.349,.1}

\journal{Medical Image Analysis}

\begin{document}

\verso{ \textit{Weng et~al.}}

\begin{frontmatter}

\title{Learning Topology-Aware Implicit Field for Unified Pulmonary Tree Modeling with Incomplete Topological Supervision}


\author[1]{Ziqiao \snm{Weng}}
\author[2,3]{Jiancheng \snm{Yang}\corref{cor1}}
\ead{jiancheng.yang@aalto.fi}
\cortext[cor1]{Corresponding author.}
\author[4]{Kangxian \snm{Xie}}
\author[5]{Bo \snm{Zhou}}
\author[1]{Weidong \snm{Cai}}

\address[1]{School of Computer Science, The University of Sydney, Sydney, NSW, Australia}
\address[2]{ELLIS Institute Finland, Espoo, Finland}
\address[3]{Aalto University, Espoo, Finland}
\address[4]{Department of Computer Science and Engineering, University of Buffalo, Buffalo, NY, USA}
\address[5]{Department of Radiology, Northwestern University, Chicago, IL, USA}


\begin{abstract}

Pulmonary trees extracted from CT images frequently exhibit topological incompleteness, such as missing or disconnected branches, which substantially degrades downstream anatomical analysis and limits the applicability of existing pulmonary tree modeling pipelines. Current approaches typically rely on dense volumetric processing, explicit graph reasoning, or generic point cloud completion priors, leading to limited efficiency, weak structural awareness, and reduced robustness under realistic structural corruption. We propose TopoField, a topology-aware implicit modeling framework that treats topology repair as a first-class modeling problem and enables unified multi-task inference for pulmonary tree analysis. TopoField represents pulmonary anatomy using sparse surface and skeleton point clouds and learns a continuous implicit field that supports topology repair without relying on complete or explicit disconnection annotations, by training on synthetically introduced structural disruptions over \textit{already} incomplete trees. Building upon the repaired implicit representation, anatomical labeling and lung segment reconstruction are jointly inferred through task-specific implicit functions within a single forward pass. Extensive experiments on the Lung3D+ dataset demonstrate that TopoField consistently improves topological completeness and achieves accurate anatomical labeling and lung segment reconstruction under challenging incomplete scenarios. We further validate TopoField on real incomplete outputs from an external segmentation model, demonstrating its applicability to realistic segmentation pipelines. Owing to its implicit formulation, TopoField attains high computational efficiency, completing all tasks in just over one second per case, highlighting its practicality for large-scale and time-sensitive clinical applications. Code and data will be available at \href{https://github.com/HINTLab/TopoField}{https://github.com/HINTLab/TopoField}

\end{abstract}


\end{frontmatter}



\section{Introduction}

Pulmonary diseases pose a major global health challenge, and computed tomography (CT)–based analysis of pulmonary airways and vasculature plays a central role in revealing disease-related morphological alterations~\citep{qin2021learning,zhang2023multi}. Accurate modeling of pulmonary tree structures underpins a wide range of clinical applications, including diagnosis of respiratory and vascular diseases, surgical planning, and image-guided interventions~\citep{fetita2004pulmonary,saji2022segmentectomy,rahaghi2016pulmonary,zhao20183d}.

In practice, pulmonary tree representations used for analysis are frequently topologically incomplete. Such defects arise from both automated CT-based extraction and manual or semi-automatic annotation, where small or low-contrast distal branches are often disconnected or omitted. Despite their diverse origins, these incomplete trees share common underlying causes related to pulmonary anatomy and imaging, including severe class imbalance, extreme thinness of peripheral branches, and complex hierarchical morphology. Consequently, topological corruption constitutes a fundamental challenge in pulmonary tree modeling and must be addressed before reliable downstream analysis, such as anatomical labeling or segment-level interpretation, can be performed.

Early topology repair methods predominantly rely on dense 3D CNNs for voxel-wise structural completion. However, most existing approaches~\citep{weng2023topology,wang2024car,carneiro2024restoring,zhou2025masked} treat repair as a post-processing step for imperfect segmentation results, often assuming fixed or highly constrained disconnection patterns or requiring costly iterative refinement. Moreover, dense volumetric modeling inherently limits computational efficiency and scalability, particularly for trees with multiple disconnections.

Generic point cloud completion methods are also relevant to this problem, as topology repair can be viewed as recovering missing structures from incomplete observations. However, conventional completion models are primarily designed to generate globally plausible object-level point sets, whereas pulmonary tree repair requires precise localization of multiple tiny, spatially dispersed disconnections along sparse tubular branches. Since these methods typically predict a fixed unordered point set rather than query-wise occupancy at arbitrary spatial locations, they are not well suited to voxel-level connectivity restoration in thin distal branches. This motivates a topology-aware implicit formulation capable of fine-grained and continuous repair.

Neural implicit functions provide an alternative paradigm by enabling efficient inference at arbitrary spatial locations, making them well suited for topology repair where missing structures must be recovered from background regions. IPGN~\citep{xie2025efficient}, the current state of the art in implicit-function-based pulmonary analysis, demonstrates strong efficiency for anatomical labeling. Nevertheless, its explicit graph-based skeleton representation relies on intact connectivity, leading to degraded robustness under disconnection scenarios. In addition, its nearest-neighbor interpolation strategy yields low-level features insufficient for fine-grained local repair, while its linear complexity with respect to input size limits scalability to large or severely corrupted trees. These limitations highlight the need for a robust, scalable, and repair-oriented implicit modeling framework.

Beyond topology repair, pulmonary tree anatomical labeling and pulmonary segment reconstruction are critical downstream tasks. Existing methods typically assume fully connected tree structures or rely on dense volumetric processing~\citep{tan2021sgnet,xie2025efficient,kuang2022makes,chen2024deep,xie2025template}, which restricts their applicability and incurs substantial computational overhead. Most importantly, current pipelines address repair, labeling, and reconstruction as separate and sequential tasks, resulting in error accumulation, inefficiency, and limited end-to-end optimization. This motivates the development of a unified framework that treats topology repair as the foundation and jointly performs multiple pulmonary tree analysis tasks within a single, coherent inference paradigm.

To this end, we propose \textbf{TopoField}, a topology-aware implicit modeling framework that treats pulmonary tree repair as a first-class problem and further unifies multiple downstream tasks within a single inference paradigm.

First, we introduce \textbf{a dedicated topology repair framework} that jointly models pulmonary tree geometry and connectivity via surface–skeleton point cloud representations, a super-point descriptor for enhanced local structural encoding, a surface-to-skeleton attention mechanism, and a unified implicit field that enables fine-grained yet scalable structural restoration.

Second, we propose \textbf{a topological repair strategy without disconnection annotations}, which allows the model to learn connectivity restoration under unknown and unlocalized disconnections by synthetically introducing additional breaks on already corrupted trees and training the network to recover these disruptions. This strategy eliminates the need for explicit breakpoint supervision while remaining consistent with real-world deployment settings, and we further validate the reliability of the resulting weak supervision.

Third, building upon the proposed repair network, we extend the framework to \textbf{a unified multi-task implicit inference scheme} that simultaneously performs pulmonary tree repair, anatomical labeling, and lung segment reconstruction within a single forward pass. While each task is realized through a task-specific implicit function, all tasks share a common multi-task implicit field for query sampling and inference. This unified design enables end-to-end optimization across tasks and efficient inference, completing all three tasks in just over one second per case, with up to nearly two orders of magnitude speedup over implicit-function baselines and around 50$\times$ speedup over sliding-window volumetric CNN baselines.Extensive experiments on the constructed Lung3D+ dataset demonstrate that TopoField not only effectively reconstructs corrupted pulmonary trees but also achieves competitive anatomical labeling and lung segment reconstruction performance, while being substantially more efficient.

\begin{figure}[tb]
    \centering
    \includegraphics[width=\linewidth,height=0.4\textheight,keepaspectratio]{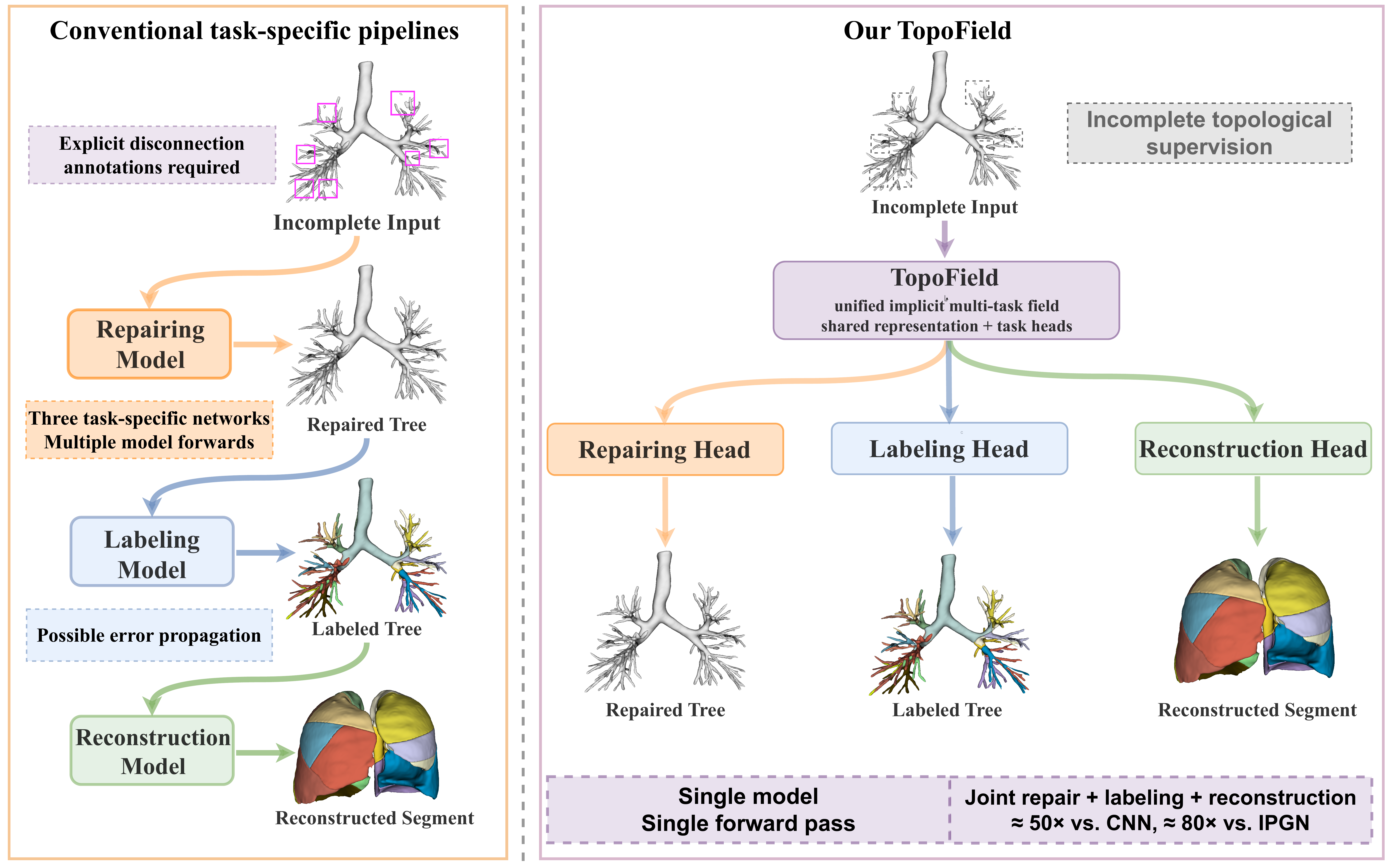}
	\caption{\textbf{Comparison between conventional task-specific pipelines and the proposed TopoField framework.} Unlike conventional pipelines that rely on separate task-specific networks, multiple model forwards, and explicit disconnection annotations for supervised repair, TopoField unifies pulmonary tree repair, anatomical labeling, and lung segment reconstruction within a single implicit multi-task field. It learns topology repair under \emph{incomplete topological supervision} without disconnection annotations and produces all task outputs in one forward pass, achieving approximately 50$\times$ and 80$\times$ faster inference than representative CNN- and IPGN-based pipelines, respectively.}
	\label{fig:fig-1}
	
\end{figure}

\textbf{Extension.} This work substantially extends our prior studies at MICCAI 2023~\citep{weng2023topology} and MIA 2025~\citep{xie2025efficient} in both the formulation of pulmonary topology repair and the scope of unified pulmonary tree analysis. Compared to our MICCAI 2023 study, which addressed topology repair using dense volumetric modeling under constrained disconnection settings, the present work generalizes repair to a continuous implicit formulation. This enables robust restoration under arbitrary and unlocalized disconnections and introduces a weakly supervised strategy that eliminates the need for explicit disconnection annotations, thereby improving applicability to realistic clinical scenarios. Beyond topology repair, this work significantly expands the task scope of IPGN~\citep{xie2025efficient}. Rather than treating anatomical labeling in isolation and assuming intact tree connectivity, we establish topology repair as the foundational task and unify repair, anatomical labeling, and lung segment reconstruction within a single implicit inference framework. All tasks are jointly optimized and inferred in one forward pass, resulting in substantially improved efficiency compared to graph-based implicit pipelines. Finally, we extend the original Lung3D dataset~\citep{xie2025template} into a comprehensive benchmark supporting multi-disconnection simulation and unified evaluation across tasks, enabling systematic assessment of repair quality, anatomical labeling, and lung segment reconstruction.

\section{Related Work}
\label{sec:related work}

\subsection{Vascular Topology Completion}

Accurate vascular segmentation is fundamental for diagnosing vascular diseases, radiotherapy planning, and surgical navigation. However, this task is intrinsically challenging due to the multi-scale, multi-orientation, and highly branching nature of vascular structures, often accompanied by frequent crossings. Traditional methods, including filtering-based enhancement~\citep{lamy2022benchmark,li2022human}, morphological operations~\citep{zana2001segmentation,tschirren2005intrathoracic}, vessel tracking~\citep{quek2002vessel,delibasis2010automatic}, and classical machine learning approaches~\citep{delibasis2010automatic,marin2010new}, rely heavily on handcrafted features, which limits their robustness and generalization ability~\citep{zhou2025masked}.

In recent years, deep learning (DL)–based approaches have substantially advanced vascular segmentation across a wide range of anatomical structures, including pulmonary airways~\citep{charbonnier2017improving,garcia2019joint,selvan2020graph,qin2019airwaynet,qin2020learning,zheng2021refined,wang2024accurate,zhang2025dmrn}, pulmonary arteries and veins~\citep{luo2023efficient,pu2023automated,chu2025deep}, retinal vessels~\citep{wu2021scs,lin2023stimulus,qin2024review,zhou2025masked}, coronary arteries~\citep{zhang2023anatomy,zeng2023imagecas,dong2023novel}, and cerebral vasculature~\citep{shi2023affinity,chen2022generative,xia20223d}. Despite these advances, achieving topologically complete vascular reconstructions remains difficult. This difficulty arises from several fundamental factors: (i) the extreme geometric complexity of vascular trees, spanning from large proximal trunks to thin distal branches with low contrast and high noise sensitivity; (ii) severe class imbalance, both between vessel and background voxels and within the vessel class itself, biasing models toward large structures; (iii) overlap-based losses such as Dice, which are dominated by large regions and fail to enforce topological consistency; and (iv) annotation imperfections, including missing or incomplete branches~\citep{zhang2023multi}. Consequently, even state-of-the-art DL models often produce fragmented and topologically inconsistent predictions.

To address these issues, prior studies have explored architectural innovations~\citep{li20213d,zhang2022progressive,zhang2023anatomy,nan2023fuzzy,zhao2023gdds,wang2024accurate}, topology-aware loss functions~\citep{yan2018joint,ke2023scale,zhang2023towards,huang2025harmonyseg}, and auxiliary supervision signals~\citep{jia2024connectivity,zhao2025skeleton2mask,he2025deep}. For instance, Zhang \textit{et al.}~\citep{zhang2022progressive} proposed a progressive learning framework that integrates anatomical context and topology modeling for coronary artery segmentation, while Shit \textit{et al.}~\citep{shit2021cldice} introduced the clDice loss to explicitly enforce centerline-level topological consistency. More recently, HarmonySeg~\citep{huang2025harmonyseg} combined vesselness-guided feature fusion with topology-preserving loss functions to mitigate scale variation and annotation incompleteness. Skeleton-supervised approaches such as Skeleton2Mask~\citep{zhao2025skeleton2mask} further reduce annotation costs while improving structural integrity.

Another line of research focuses on post-processing refinement. Traditional methods rely on heuristic techniques such as conditional random fields~\citep{fu2016retinal}, morphological operations~\citep{fraz2012ensemble}, or gradient-based extrapolation~\citep{zhang2018reconnection}, which are non-learnable and difficult to optimize. Learning-based refinement strategies have therefore been proposed~\citep{yan2018three,nadeem2020ct,wu2020nfn+,wang2024car,yang2024multi,zhou2025masked}. These include cascaded or multi-stage frameworks that progressively recover peripheral branches and repair disconnections. Some approaches synthesize artificial disconnections to train reconnection models~\citep{weng2023topology,weng2024efficient,wang2024car,carneiro2024restoring,carneiro2024plug}, though they often rely on fixed disconnection assumptions or computationally expensive iterative refinement. MaskVSC~\citep{zhou2025masked} integrates simulated connectivity learning and a Vision GNN–based refinement stage, but is restricted to 2D retinal imagery.

Despite these efforts, most existing methods remain based on dense voxel-wise CNNs, which are memory-intensive and computationally expensive, particularly for high-resolution 3D data. Such models often operate on local patches, resulting in limited scalability and inconsistent global topology. In contrast, our framework reformulates vascular topology completion as an implicit inference problem over sparse point-cloud representations, enabling efficient full-volume repair within a continuous space. Moreover, our weakly supervised strategy eliminates the need for explicit disconnection annotations. Beyond topology completion, the proposed approach unifies vascular repair, anatomical labeling, and lung segment reconstruction within a single multi-task implicit framework, achieving both topological completeness and semantic consistency.

\subsection{Pulmonary Structure Semantic Segmentation}

Accurate semantic modeling of pulmonary structures, including airways, vessels, and lung segments, is fundamental for anatomical understanding and clinical applications such as disease diagnosis, surgical planning, and navigation. Multi-class pulmonary tree segmentation aims to assign anatomically meaningful labels to each branch but remains challenging due to complex tree topology and unreliable distal structures. Pulmonary segments, defined by segmental bronchi and arteries and bounded by intersegmental veins, constitute anatomically and functionally independent units, making their accurate delineation particularly important for segmentectomy planning~\citep{xie2025template}. Semantic segmentation further enables downstream tasks such as lesion localization and functional analysis~\citep{tan2021sgnet}.

Existing studies have explored pulmonary tree and segment labeling using structure-aware representations~\citep{nadeem2020anatomical,tan2021sgnet,yu2022tnn,huang2024bcnet,xie2024structure,chau2024branchlabelnet}. SGNet~\citep{tan2021sgnet} integrates airway segmentation and semantic labeling within a graph-based framework, leveraging anatomical priors to improve topological consistency. Subsequent works~\citep{kuang2022makes,xie2025efficient,xie2025template} further extended semantic segmentation to full lung segments and multi-structure annotation. Notably, IPGN~\citep{xie2025efficient} represents pulmonary trees as sparse point clouds and skeleton graphs for implicit semantic labeling, but requires topology-complete inputs and multi-stage training, limiting robustness and efficiency.

Implicit representation–based methods such as ImPulSe~\citep{kuang2022makes} and deformable template networks~\citep{xie2025template} have improved pulmonary segment reconstruction accuracy. However, these approaches generally assume fully segmented and connectivity-complete trees, which is often unrealistic in clinical CT data.

In contrast, TopoField differs fundamentally from prior work by explicitly addressing tree incompleteness before semantic modeling. It first repairs disconnected pulmonary structures, then unifies connectivity completion, tree semantic labeling, and lung segment reconstruction within a single implicit framework. This design avoids reliance on topology-complete inputs and multi-stage pipelines, resulting in substantially improved inference efficiency and robustness.

\section{Problem Formulation}

\subsection{Task Description}

We consider three closely related tasks—\textit{pulmonary tree repair}, \textit{anatomical labeling}, and \textit{lung segment reconstruction}, which together enable comprehensive and anatomically consistent modeling of the pulmonary system. These tasks are inherently interdependent: reliable semantic labeling and segment-level interpretation critically depend on a topologically complete pulmonary tree representation.

\paragraph{Pulmonary tree repair}
Pulmonary tree representations used in practice are frequently affected by topological disconnections, arising from both automated CT-based extraction and manual or semi-automatic annotation, where distal and low-contrast branches are often disconnected or omitted. Despite their different sources, such defects are consistently rooted in the intrinsic challenges of pulmonary anatomy and imaging, including complex tubular morphology and severe foreground–background class imbalance. These disconnections substantially impair downstream analyses and therefore necessitate a dedicated topology repair step. The central challenge of pulmonary tree repair lies in accurately identifying and restoring missing distal branches from large background regions, while ensuring that the reconstructed structures remain anatomically plausible and topologically consistent with the original tree morphology.

\paragraph{Anatomical labeling}
The anatomical labeling task assigns each pulmonary airway, artery, and vein branch to its anatomically defined segmental class. Following~\citep{kuang2022makes,xie2025efficient}, the lung is partitioned into 18 anatomical segments (8 in the left lung and 10 in the right), and the pulmonary tree is further annotated into 19 semantic classes, including 18 intra-lobar (\textit{peripheral}) classes and one extra-lobar (\textit{central}) class. This task is particularly sensitive to topological integrity, as critical skeleton points defining inter-class boundaries and global hierarchy are sparse and easily disrupted by disconnections. Consequently, labeling accuracy strongly depends on a topologically complete tree, while standard voxel-wise metrics often underrepresent errors at these anatomically critical locations.

\paragraph{Lung segment reconstruction}
The lung segment reconstruction task aims to recover the complete 3D structure of each anatomical segment based on its associated segmental bronchi, arteries, and intersegmental veins. Beyond voxel-level accuracy, this task emphasizes anatomically correct inter-segment boundaries and structural coherence with the underlying pulmonary trees. Accurate reconstruction is therefore challenging when tree connectivity is incomplete or inconsistent.

\paragraph{Unified formulation}
Motivated by the above observations, we formulate pulmonary tree repair as a foundational task and propose a unified framework in which anatomical labeling and lung segment reconstruction are performed on the repaired topology. Rather than treating these tasks as independent or sequential post-processing steps, our approach jointly models geometry, topology, and semantics within a shared implicit representation, enabling consistent and efficient inference across all three tasks within a single framework. The technical details of the proposed unified implicit modeling approach are presented in Section~\ref{sec:method}.

\begin{figure}[tb]
    \centering
	\includegraphics[width=\linewidth]{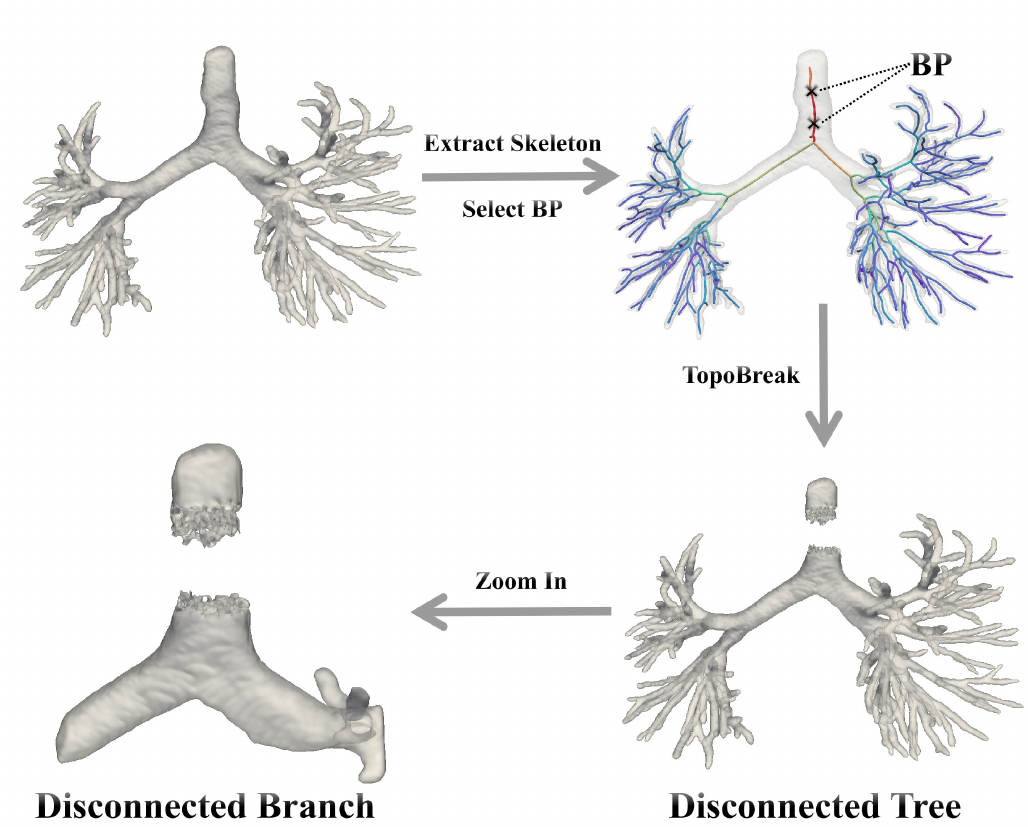}
	\caption{\textbf{Overview of the proposed TopoBreak strategy.}. Starting from a complete pulmonary tree mask, the skeleton is extracted and two breakpoints (BP) are selected along a branch. Guided by the skeleton and branch radius, TopoBreak removes the segment between these points to create realistic topological disconnections, as shown in the close-up view.}
	\label{fig:fig-2}
	
\end{figure}

\subsection{Topology-Aware Mask Strategy}
\label{sec:mask}

In medical imaging, data scarcity often limits model robustness and generalization. To alleviate this issue and enhance the diversity of training data, we propose a topology-aware masking strategy, termed \textbf{TopoBreak}, which leverages the morphological characteristics of tubular structures to simulate realistic structural corruptions. Specifically, TopoBreak randomly removes small continuous segments of pulmonary branches, thereby generating synthetic disconnections or missing regions that mimic common artifacts in predicted tree segmentations, as illustrated in Fig.~\ref{fig:fig-2}.

The corruption generation pipeline proceeds as follows. For each annotated pulmonary tree, we first convert the corresponding CT volume into a binary mask, where foreground voxels (value 1) represent tree structures and background voxels (value 0) correspond to non-tree regions. We then extract the skeleton of each tree using the VesselVio toolkit~\citep{bumgarner2022vesselvio}, which employs the classical thinning algorithm of Lee \emph{et al.}~\citep{lee1994building}. VesselVio further provides a graph-based representation of the skeleton, where each edge corresponds to an individual branch and encodes both spatial and morphological attributes, including the 3D coordinates of skeleton points, as well as the branch radius, length, and volume.

Based on this graph, we randomly select a vessel branch and identify two non-overlapping breakpoints along its skeleton. The segment between these two points is then removed using morphological operations guided by the branch radius, producing a realistic topological disconnection. To avoid overly smooth cross-sections that deviate from real-world break patterns, we do not completely erase all voxels within the selected segment. Instead, we adopt a spatially adaptive removal scheme: voxels near the two breakpoint cross-sections are more likely to be retained, while those closer to the segment’s midpoint have progressively lower retention probabilities, approaching zero at the center. This results in discontinuities with irregular, anatomically plausible boundaries. To generate multi-disconnection samples, the procedure can be repeated multiple times on the same pulmonary tree.

\subsection{Dataset \label{sec:dataset}}

We constructed the \textbf{Lung3D+} dataset based on the publicly available PTR dataset~\citep{weng2023topology} and Lung3D dataset~\cite{xie2025template}. The Lung3D+ dataset comprises 799 subjects collected from multiple medical centers, with radiologist-provided annotations of the pulmonary airways, arteries, and veins. The annotations cover 19 semantic classes, where peripheral branches in the left lung correspond to labels 1–10, those in the right lung to labels 11–18, and the central tree structures outside the lungs to label 19. Each subject is represented by a 3D volume of size (N, 512, 512), where 512×512 corresponds to the in-plane resolution of each slice and N denotes the number of slices, ranging from 181 to 798. For unbiased evaluation, the dataset was split into 70\% for training (559 subjects), 10\% for validation (80 subjects), and 20\% for testing (160 subjects).

\begin{figure}[tb]
    \centering
	\includegraphics[width=\linewidth]{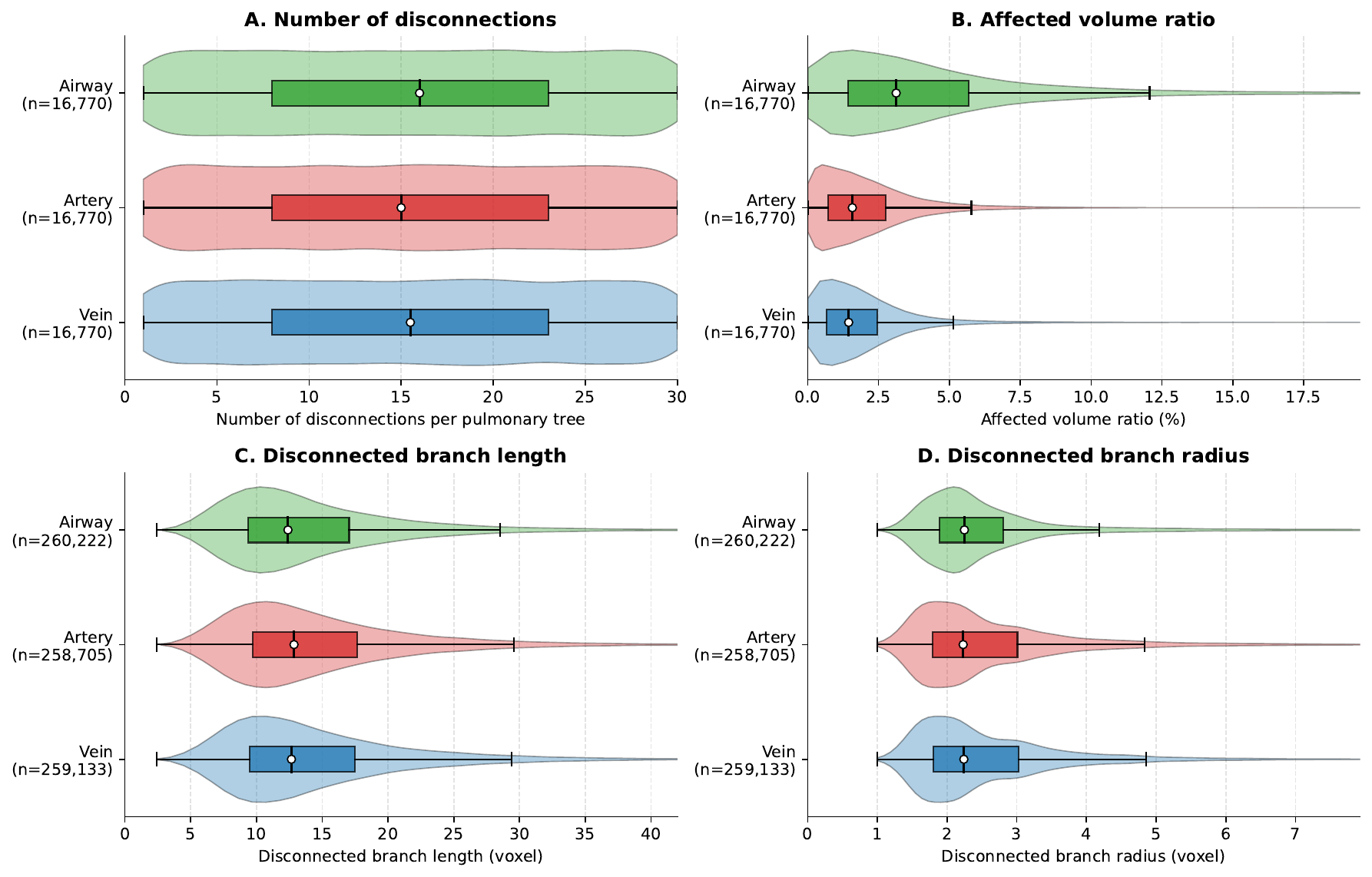}
	\caption{\textbf{Quantitative characterization of synthesized disconnections in Lung3D+}. The distributions of sample-level and disconnection-level statistics are summarized for the airway, artery, and vein training datasets. (A) Number of synthesized disconnections per pulmonary tree. (B) Affected volume ratio, computed as the number of disconnected points divided by the number of volume points. (C) Disconnected branch length in voxels. (D) Disconnected branch radius in voxels. For panels B–D, the x-axis is truncated at the 99th percentile across the three datasets.}
	\label{fig:lung3d_corruption_statistics}
	
\end{figure}

The Lung3D+ dataset includes not only the complete pulmonary trees and lung segments but also corrupted tree data synthesized using TopoBreak for the task of topological repair. Specifically, for each training subject, we generated 30 corrupted versions of the airways, arteries, and veins with varying levels of disconnection. For the validation and test sets, we generated three corrupted versions per subject. This process resulted in a total of 16,770 training samples, 240 validation samples, and 480 test samples.
For each synthesized disconnected tree data, we generate 1 to 30 disconnections across different branches, yielding multiple disconnected trees. Figure~\ref{fig:lung3d_corruption_statistics} further summarizes the distributions of the synthesized disconnections across the airway, artery, and vein training datasets. Our ultimate objective is to repair these disconnections and simultaneously perform semantic labeling across the complete pulmonary tree and semantic reconstruction of lung segments. Our ultimate objective is to repair these disconnections and simultaneously perform semantic labeling across the complete pulmonary tree and semantic reconstruction of lung segments.

\begin{figure}[tb]
    \centering
	\includegraphics[width=\linewidth]{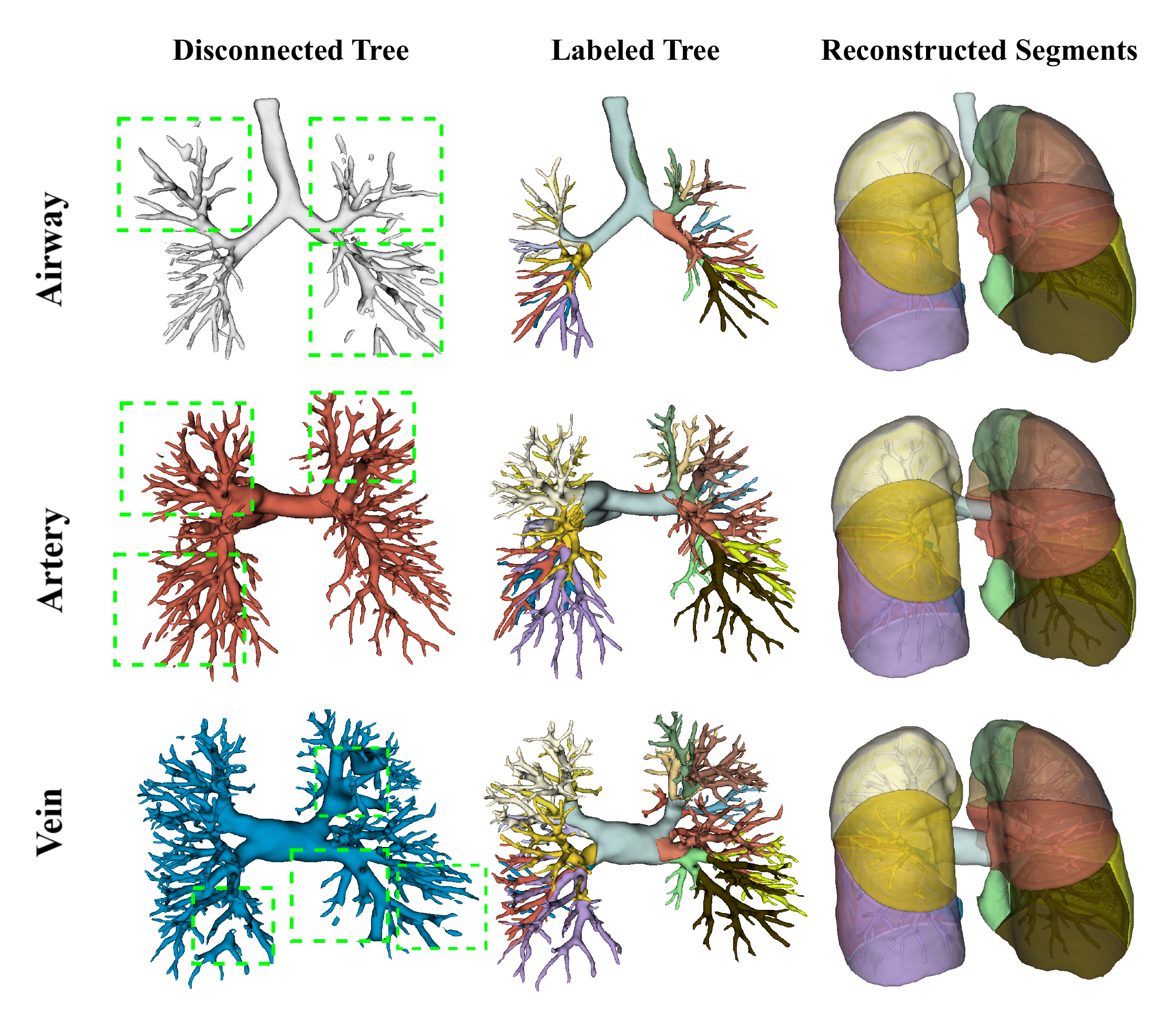}
	\caption{\textbf{An overview of the Lung3D+ dataset}. The dataset contains both complete and simulated disconnected pulmonary trees across the airways, arteries, and veins. For clarity, we illustrate the disconnected input trees alongside their corresponding complete labeled trees and lung segments. All three types of trees belong to the same subject and share identical segmental anatomy. In the corrupted binary trees, disconnection sites are highlighted and magnified with green boxes, indicating regions of structural disruption requiring repair.}
	\label{fig:fig-3}
	
\end{figure}

An overview of the Lung3D+ dataset, including complete/corrupted pulmonary trees and lung segments, is illustrated in Figure~\ref{fig:fig-3}.

\subsection{Evaluation Metrics \label{sec:evaluation}}

\subsubsection{Repair Metrics}

Repairing pulmonary disconnections is particularly challenging, as it requires the accurate identification of fine-grained local disconnections and their anatomically plausible repair while maintaining the continuity of tubular structures. At the same time, the repair must effectively discriminate between foreground and background to avoid introducing spurious branches and thereby reduce false positives. To comprehensively evaluate repair performance, we design a repair metric system comprising four complementary measures. Let the set of ground-truth disconnected components be denoted by $\{G_1, G_2, \dots, G_N\}$ and the set of predicted disconnected components by $\{P_1, P_2, \dots, P_M\}$, both obtained through connected-component analysis.

\paragraph{Containment F1 (CF1)}  
This metric focuses on instance-level detection accuracy and primarily evaluates the model's capability to successfully identify each disconnected component, regardless of detailed shape fidelity. Specifically, for each ground-truth disconnected component $G_i$, if the overlap ratio 

\begin{equation}
\scalebox{0.9}{$\displaystyle
\frac{|G_i \cap \bigcup_{j=1}^{M} P_j|}{|G_i|} > 0.5,
$}
\end{equation}

then $G_i$ is considered successfully detected (recall). Similarly, for each predicted component $P_j$, if
\begin{equation}
\scalebox{0.9}{$\displaystyle
\frac{|P_j \cap \bigcup_{i=1}^{N} G_i|}{|P_j|} > 0.5,
$}
\end{equation}
then $P_j$ is considered a correct detection (precision). The CF1 score is obtained by combining precision and recall through their harmonic mean.

\paragraph{Dice Matching F1 (DMF1)}  
This metric reflects high-level object-level detection quality in pulmonary reconstruction. It requires both detection and sufficient shape overlap, thereby providing a stricter evaluation of reconstruction fidelity. For each ground-truth component $G_i$, we compute the Dice coefficient with every predicted component:
\begin{equation}
\scalebox{0.9}{$\displaystyle
\text{Dice}(G_i, P_j) = \frac{2|G_i \cap P_j|}{|G_i| + |P_j|}, \quad j=1,\dots,M,
$}
\end{equation}
and take the maximum. If $\max_j \text{Dice}(G_i, P_j) > 0.5$, then $G_i$ is considered successfully detected (recall). Conversely, for each predicted component $P_j$, if $\max_i \text{Dice}(G_i, P_j) > 0.5$, it is considered a correct detection (precision). The DMF1 score is then computed as the harmonic mean of precision and recall.

\paragraph{Global Dice (GDice)}  
To assess disconnection reconstruction quality at the volume level, we compute the Dice coefficient between the union of all predicted and all ground-truth components:
\begin{equation}
\scalebox{0.9}{$\displaystyle
\text{GDice} = \frac{2 \left|\left(\bigcup_{i=1}^{N} G_i\right) \cap \left(\bigcup_{j=1}^{M} P_j\right)\right|}
{\left|\bigcup_{i=1}^{N} G_i\right| + \left|\bigcup_{j=1}^{M} P_j\right|}.
$}
\end{equation}

\paragraph{Number of Connected Components (NCC)}  
This metric quantifies structural completeness by computing the number of connected components present in the reconstructed pulmonary tree. The reconstructed tree is defined as the union of the original disconnected pulmonary tree and all predicted components. A smaller NCC indicates better connectivity, with the ideal case being $\text{NCC}=1$, representing a fully connected tree structure.

\subsubsection{Labeling and Reconstruction Metrics}

Pulmonary tree labeling requires accurate discrimination of fine-grained local structures while preserving global anatomical coherence across categories. To evaluate labeling performance over the 19 anatomical classes, we adopt a comprehensive metric suite. Given the severe class imbalance, most notably the dominance of the central tree class, we follow~\citep{xie2025efficient} and use the \textit{micro-averaged Dice coefficient} as the primary metric.

To further assess performance at different structural scales, we report two complementary Dice-based metrics.
Voxel-level Dice measures voxel-wise agreement between predictions and ground truth for each class, reflecting overall classification accuracy. Skeleton-level Dice evaluates labeling accuracy on the extracted tree skeletons, focusing on structurally critical locations that define global topology.

Notably, voxel-level Dice is computed on the disconnected pulmonary tree provided as input to the model, whereas skeleton-level Dice is evaluated on skeletons extracted from the complete ground-truth tree. This design is intentional: voxel-level Dice captures classification consistency under incomplete structural input, while skeleton-level Dice assesses anatomical correctness along the principal structural axes. Evaluating skeleton Dice on disconnected trees is unreliable, as disconnections often introduce spurious local cliques that do not correspond to true anatomical skeletons.

For lung segment reconstruction, we use voxel-level Dice to evaluate all 18 anatomical segments over the entire lung volume, providing a holistic assessment of intra-segment consistency and inter-segment boundary accuracy.

\section{Method \label{sec:method}}
\begin{figure*}[tb]
    \centering
	\includegraphics[width=\linewidth]{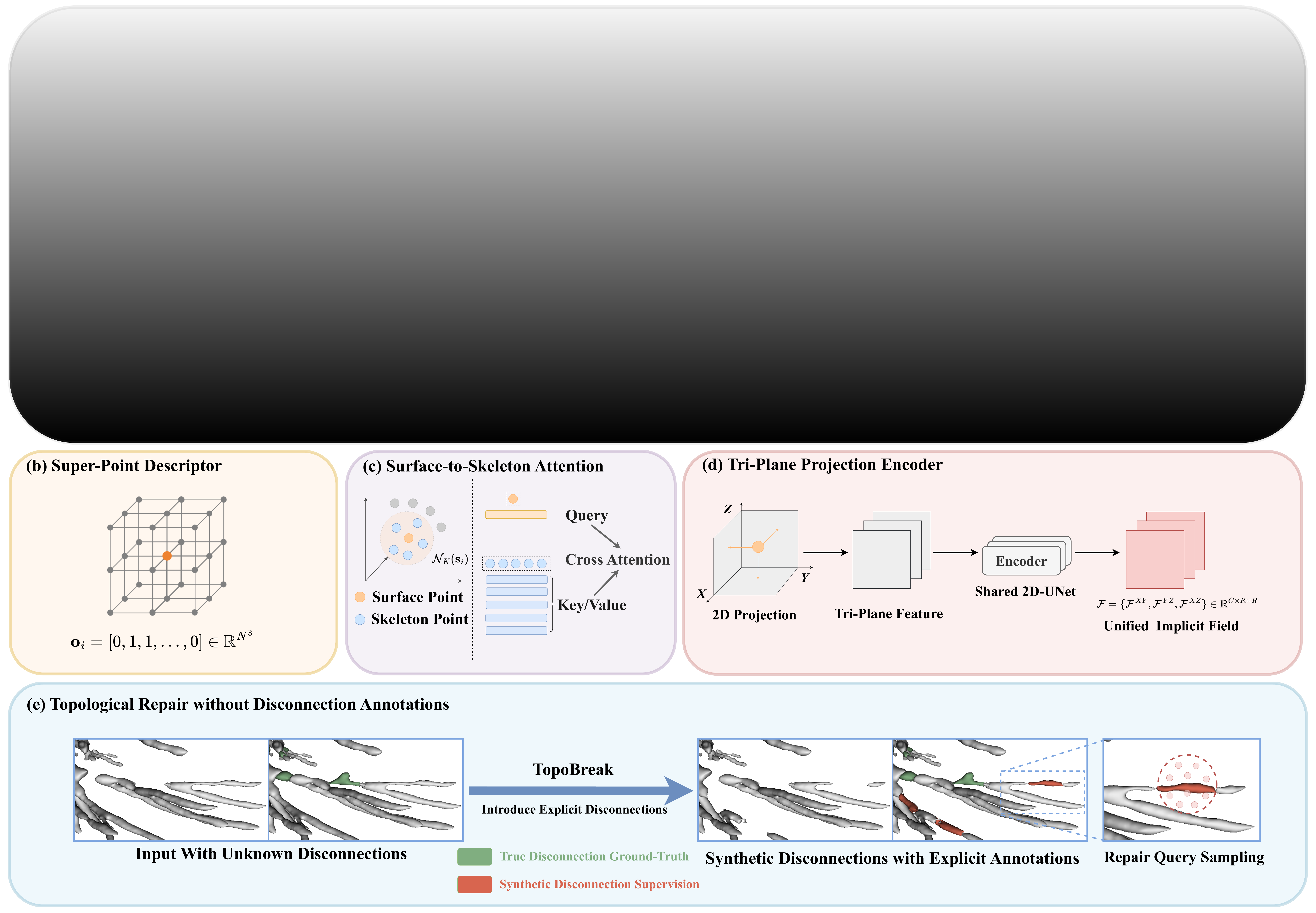}
	\caption{\textbf{Overview of our TopoField framework.} 
(a) The incomplete pulmonary tree is represented using both a surface point cloud ($\mathcal{S}$) and a skeleton point cloud ($\mathcal{K}$), which are respectively encoded by point-based networks. Surface features are further enriched by a super-point descriptor, while skeleton features capture the global topological structure. The two representations are fused via the proposed Surface-to-Skeleton Attention (SSA) module, where surface points act as queries and skeleton points as keys/values. The fused features are then projected onto three orthogonal planes to generate tri-plane feature maps, which are jointly processed by a shared 2D U-Net encoder to construct a unified multi-task implicit field. Task-specific implicit functions subsequently enable efficient inference on sampled query points during training and dense full-volume queries during inference, supporting pulmonary tree repair, anatomical labeling, and lung segment reconstruction simultaneously. (b)–(d) illustrate the detailed designs of the corresponding modules. (e) illustrates our \emph{Topological Repair without Disconnection Annotations}, which implicitly restores missing connectivity without requiring explicit disconnection annotations.
}
	\label{fig:fig-4}
\end{figure*}

In this section, we present a unified framework for \textbf{pulmonary tree topology repair}, which is further extended to support \textbf{anatomical labeling} and \textbf{lung segment reconstruction}, as illustrated in Figure~\ref{fig:fig-4}. We begin in Section~4.1 by formally defining the three target tasks and their respective learning objectives. Section~4.2 introduces the core of our approach, detailing how the proposed framework achieves robust repair under incomplete topological supervision. Specifically, the repair capability is enabled by three key components. First, we propose a \textit{pulmonary tree modeling} strategy that jointly encodes surface and skeletal structures, effectively integrating geometric and topological cues. Second, we construct a \textit{unified implicit field} that embeds the encoded tree representation into a continuous feature space, enabling query inference at arbitrary spatial locations. Third, we develop a \textit{weakly supervised training} strategy that facilitates disconnection repair without requiring explicit annotations of true breakpoints. Building upon this repair-oriented formulation, Section~4.3 extends the framework to a multi-task setting, where anatomical labeling and lung segment reconstruction are jointly inferred by querying the shared multi-task implicit field. Finally, Section~4.4 describes the overall training objective that unifies all tasks within a single optimization framework.

\subsection{Problem Definition}

We consider three interrelated tasks: \emph{pulmonary tree repair}, \emph{anatomical labeling}, and \emph{lung segment reconstruction}.

\paragraph{Pulmonary tree repair and anatomical labeling}
Given a corrupted pulmonary tree volume
\begin{equation}
\scalebox{0.95}{$\displaystyle
\mathcal{T} \in \{0,1\}^{X \times Y \times Z},
$}
\end{equation}
where foreground voxels represent incomplete airway or vascular structures, the objective is to recover a topologically complete tree and assign each foreground voxel an anatomical label. Formally, we aim to predict
\begin{equation}
\scalebox{0.95}{$\displaystyle
f_l(\mathcal{T}) \;=\; \hat{\mathcal{Y}}, 
\quad \hat{\mathcal{Y}} \in \{0,1,\dots,19\}^{X \times Y \times Z},
$}
\end{equation}
where label $0$ denotes background, labels $1$--$18$ correspond to peripheral branches within the 18 lung segments, and label $19$ represents central root structures outside segmental regions.  
This formulation jointly captures both \emph{topological repair} and \emph{semantic labeling} within a unified prediction space.

\paragraph{Lung segment reconstruction}
Given a complete lung volume
\begin{equation}
\scalebox{0.95}{$\displaystyle
\mathcal{L} \in \{0,1\}^{X \times Y \times Z},
$}
\end{equation}
the goal is to reconstruct a voxel-wise segmentation into 18 anatomical lung segments:
\begin{equation}
\scalebox{0.95}{$\displaystyle
f_s(\mathcal{L}) \;=\; \hat{\mathcal{S}}, 
\quad \hat{\mathcal{S}} \in \{1,\dots,18\}^{X \times Y \times Z}.
$}
\end{equation}
This task emphasizes anatomically consistent delineation of inter-segment boundaries, guided by the underlying pulmonary tree structure.

\subsection{Robust Topological Repair}

\paragraph{Pulmonary Tree Modeling}
To efficiently represent complex pulmonary structures while preserving both geometry and topology, we decompose each corrupted tree $\mathcal{T}$ into two complementary point clouds: a surface representation and a skeletal representation.

The surface point cloud
\begin{equation}
\scalebox{0.95}{$\displaystyle
\mathcal{S} = \{\mathbf{s}_i\}_{i=1}^{N_s} \subset \mathbb{R}^3
$}
\end{equation}
is obtained by extracting boundary voxels from $\mathcal{T}$ and uniformly downsampling them to $N_s$ points.  
To encode global topology without relying on fragile graph connectivity, we extract a skeleton point cloud
\begin{equation}
\scalebox{0.95}{$\displaystyle
\mathcal{K} = \{\mathbf{k}_j\}_{j=1}^{N_k} \subset \mathbb{R}^3
$}
\end{equation}
using a thinning algorithm~\cite{lee1994building}, followed by uniform downsampling or padding.

Both point sets are processed by independent Point-Voxel CNN encoders~\cite{liu2019point}, producing feature embeddings
\begin{equation}
\scalebox{0.95}{$\displaystyle
\Phi_s \in \mathbb{R}^{N_s \times d}, 
\qquad
\Phi_k \in \mathbb{R}^{N_k \times d}.
$}
\end{equation}
This dual representation balances fine-grained geometric detail and global structural context, forming the basis for topology-aware fusion.

To enrich sparse surface representations with local structural context, we introduce a \emph{super-point descriptor}. For each surface point $\mathbf{s}_i$, a cubic neighborhood of radius $r$ is defined, and voxel occupancy within this region is flattened into a local descriptor, capturing continuity and density patterns that are critical for topology inference.

To integrate geometric and topological cues, we propose a \emph{Surface-to-Skeleton Attention (SSA)} module. For each surface point $\mathbf{s}_i$, we identify its $K$ nearest skeleton neighbors, denoted as $\mathcal{N}_K(\mathbf{s}_i)$, and cross-attention is applied with surface features as queries and skeleton features as keys and values. The fused surface feature is obtained via residual aggregation:
\begin{equation}
\scalebox{0.95}{$\displaystyle
\hat{\phi}^s_i \;=\; \phi^s_i \;+\; 
\mathrm{Attn}\!\left(\phi^s_i,\{\phi^k_j\}_{j \in \mathcal{N}_K(\mathbf{s}_i)}\right).
$}
\end{equation}
This design injects topology-aware information while preserving geometric fidelity and avoids the instability of graph-based message passing under severe disconnections.

\paragraph{Unified Implicit Field}
Inspired by \citet{peng2020convolutional}, the fused surface features are projected onto three orthogonal planes (XY, YZ, XZ) with a resolution of $R$ and refined using a shared 2D U-Net encoder, yielding a tri-plane implicit field
\begin{equation}
\scalebox{0.95}{$\displaystyle
\mathcal{F} = \{\mathcal{F}^{XY}, \mathcal{F}^{YZ}, \mathcal{F}^{XZ}\}
\in \mathbb{R}^{C \times R \times R}.
$}
\end{equation}
This encoded field provides a compact, continuous representation of pulmonary geometry and topology:
\begin{equation}
\scalebox{0.95}{$\displaystyle
\mathcal{F}
=
\mathrm{UNet}\!\left(
\mathrm{Proj}\!\left(
\mathrm{SSA}(\Phi_s,\Phi_k) \mid \mathcal{S}
\right)
\right).
$}
\end{equation}

\paragraph{Repair Query Sampling}
To enable voxel-wise prediction over the continuous implicit field, we employ a repair-specific implicit function $f_r(\cdot)$ parameterized by a lightweight multilayer perceptron (MLP). 
For any query point $\mathbf{q} \in \mathbb{R}^3$, its local features are obtained via trilinear interpolation from the tri-plane implicit field $\mathcal{F}$, yielding
$\{\mathbf{f}^{XY}_q, \mathbf{f}^{YZ}_q, \mathbf{f}^{XZ}_q\}$.
To preserve spatial awareness, a positional encoding $g(\mathbf{q}) \in \mathbb{R}^D$ is concatenated with the sampled features, forming the query embedding
\begin{equation}
\scalebox{0.95}{$\displaystyle
\mathbf{h}_q =
\big[
\mathbf{f}^{XY}_q;
\mathbf{f}^{YZ}_q;
\mathbf{f}^{XZ}_q;
g(\mathbf{q})
\big],
$}
\end{equation}
which is subsequently decoded by $f_r(\mathbf{h}_q)$ to predict voxel occupancy indicating whether the queried location should be reconnected.

Query sampling plays a critical role in effective topology repair. 
For the \textbf{repair task}, we explicitly exploit the locations of synthetically introduced disconnections during training.
At each iteration, one disconnection is randomly selected, and its two skeleton endpoints are identified.
Along the line segment connecting these endpoints, we sample $Q_r \cdot p$ query points within a fixed-radius neighborhood, where $Q_r$ denotes the total number of repair queries and $p \in (0,1)$ controls the proportion of samples concentrated within the missing region.
The remaining $Q_r(1-p)$ queries are uniformly sampled from background voxels to regularize the learning process and suppress false-positive reconnections.

\paragraph{Topological Repair without Disconnection Annotations}

In practical clinical settings, the precise locations and extents of disconnections within corrupted pulmonary trees are typically unknown, rendering fully supervised topological repair infeasible.
To overcome this limitation, we propose a weakly supervised training strategy that enables effective structural restoration without requiring explicit disconnection annotations.

The proposed strategy leverages the localized nature of repair query sampling.
Given a corrupted pulmonary tree volume $\mathcal{T}$ with unknown disconnections, we apply the proposed $\mathrm{TopoBreak}(\cdot)$ operator to synthetically introduce an additional, explicit disconnection.
This yields an augmented tree $\tilde{\mathcal{T}} = \mathrm{TopoBreak}(\mathcal{T})$, while the original tree $\mathcal{T}$ is treated as a weak supervisory target.
The repair network is then trained to reconstruct $\mathcal{T}$ from $\tilde{\mathcal{T}}$, enforcing the recovery of structural continuity at the newly introduced breakpoints while preserving the global topology of the originally corrupted tree.

To ensure that the synthetic supervision remains structurally meaningful, $\mathrm{TopoBreak}$ is applied only to branches whose skeletons exceed a predefined node threshold.
Repair query sampling remains identical to the strategy introduced earlier, with queries concentrated around the synthetically generated disconnection regions to promote focused learning of missing connectivity.

Although this weakly supervised formulation removes the need for explicit disconnection annotations, it may introduce limited supervision noise.
In particular, repair queries sampled from synthetic disconnection regions may occasionally overlap with true disconnections already present in $\mathcal{T}$.
In such cases, regions that genuinely require repair may be mistakenly treated as intact, effectively converting true positives into false negatives.

To quantify the reliability of the resulting supervision, we estimate its expected accuracy under a simplified probabilistic assumption.
Let $\Omega_T$ denote the voxel support of the corrupted pulmonary tree $\mathcal{T}$, and let $\rho_d \in [0,1]$ denote the proportion of voxels corresponding to true disconnections.
Repair query centers are sampled uniformly from $\Omega_T$, with each center defining a fixed-radius sampling region.
The induced repair query sampling space is denoted as $\Omega_q$, with volume $|\Omega_q|$.
Under this formulation, the expected accuracy of the synthetic supervision can be approximated as
\begin{equation}
\label{eq:weak-supervision-accuracy}
\scalebox{0.95}{$\displaystyle
\mathbb{E}[\mathrm{Acc}] = 1 - \frac{|\Omega_T| \cdot \rho_d}{|\Omega_q|}.
$}
\end{equation}
Since $|\Omega_q| \gg |\Omega_T|$ in practice and $\rho_d$ is typically small, the expected supervision accuracy remains high.
On the Lung3D+ dataset, we empirically estimate this accuracy to be 99.34\%, indicating that the proposed weakly supervised strategy provides reliable supervisory signals in realistic scenarios.

During inference, the network operates identically to the fully supervised setting, directly processing corrupted pulmonary trees with unknown disconnections.
Despite being trained without explicit disconnection annotations, the model generalizes effectively to unseen disconnection patterns, enabling robust and reliable topological repair in real-world applications.

\subsection{Unified Multi-Task Implicit Inference}

Building upon the proposed topology repair framework, we further extend our approach to jointly support \emph{pulmonary tree repair}, \emph{anatomical labeling}, and \emph{lung segment reconstruction} within a unified implicit inference paradigm. Notably, this multi-task formulation is not intended to enforce mutual improvement among the three tasks, but rather to establish a unified repair-centered implicit inference framework for pulmonary tree analysis. Once a corrupted pulmonary tree is encoded into the shared implicit feature field, multiple task-specific implicit functions can be applied to enable simultaneous multi-task prediction.

In addition to the repair function, we introduce two implicit functions, $f_{\text{l}}(\cdot)$ and $f_{\text{s}}(\cdot)$, each parameterized by a MLP. All implicit functions operate on the same latent implicit field, while being optimized independently to accommodate the distinct objectives of anatomical labeling and lung segment reconstruction.

For the \textbf{anatomical labeling} task, $Q_l$ query points are uniformly sampled from the foreground voxels of the corrupted pulmonary tree.
Each query is supervised with its corresponding anatomical class label obtained from ground-truth annotations.
For the \textbf{lung segment reconstruction} task, $Q_s$ query points are uniformly sampled from the left and right lung regions, and the network predicts their associated segmental classes.

Although all tasks share a common latent representation, each is realized through a dedicated implicit function.
Specifically, the repair function predicts a binary occupancy value $\hat{o}_r \in \{0,1\}$, where $\hat{o}_r = 1$ denotes voxels that should be reconnected to restore structural continuity.
In contrast, the labeling and reconstruction functions output categorical predictions $\hat{y}_l \in \{1,\dots,19\}$ and $\hat{y}_s \in \{1,\dots,18\}$, corresponding to pulmonary tree anatomical classes and lung segment categories, respectively.
This design enables efficient and coherent inference of multiple interdependent tasks within a single forward pass.

\paragraph{Inference}
During inference, given a corrupted pulmonary tree with unknown disconnection locations, we first extract its downsampled surface and skeleton representations and encode them into a unified multi-task implicit feature field.
Task-specific query strategies are then applied to enable full-volume prediction within this shared field.

For pulmonary tree repair and anatomical labeling, all background voxels are treated as \textit{repair queries}, while all foreground tree voxels are treated as \textit{labeling queries}, enabling simultaneous inference of structural restoration and semantic classification across the entire volume.
The predicted reconnections are subsequently fused with the input corrupted tree to produce a topologically complete structure, upon which the anatomical labeling function assigns semantic classes to each voxel, yielding the final labeled pulmonary tree.

Evaluation metrics are computed only over the originally observed regions, as ground-truth annotations are available exclusively for these voxels.
Predictions in newly reconstructed regions are instead provided as reliable candidates for manual verification or semi-automatic annotation in clinical practice.

For lung segment reconstruction, all voxels within each lung lobe are treated as \textit{reconstruction queries}, allowing the model to generate complete and anatomically consistent segmentations of all 18 lung segments within the same unified inference framework.

\subsection{Training Objective}

Our framework is optimized in an end-to-end manner using a unified multi-task objective that jointly supervises \emph{tree repair}, \emph{anatomical labeling}, and \emph{lung segment reconstruction}. Each task contributes a complementary learning signal to promote both local continuity restoration and global anatomical consistency.

\paragraph{Repair Loss.}  
To ensure precise occupancy recovery and topological coherence, we employ a weighted combination of Binary Cross-Entropy (BCE) and Dice losses:
\begin{equation}
\scalebox{0.9}{$\displaystyle
\mathcal{L}_{\mathrm{repair}} 
= \lambda_{\mathrm{bce}}\, \mathcal{L}_{\mathrm{bce}}(\hat{o}, o) 
+ \lambda_{\mathrm{dice}}\, \mathcal{L}_{\mathrm{dice}}(\hat{o}, o),
$}
\end{equation}
where $\hat{o}$ and $o$ denote the predicted and ground-truth occupancy labels, respectively. This formulation balances voxel-level accuracy and region-level structural consistency.

\paragraph{Labeling and Reconstruction Loss.}  
For anatomical labeling and lung segment reconstruction, categorical Cross-Entropy (CE) loss is used to supervise semantic predictions:
\begin{equation}
\scalebox{0.9}{$\displaystyle
\mathcal{L}_{\mathrm{label}} = \mathcal{L}_{\mathrm{CE}}(\hat{y}_l, y_l), 
\quad 
\mathcal{L}_{\mathrm{recon}} = \mathcal{L}_{\mathrm{CE}}(\hat{y}_s, y_s),
$}
\end{equation}
where $\hat{y}_l$ and $\hat{y}_s$ are the predicted class probabilities for labeling and reconstruction, and $y_l$, $y_s$ are their corresponding ground-truth labels.

\paragraph{Overall Objective.}  
The final loss integrates all task-specific components:
\begin{equation}
\scalebox{0.9}{$\displaystyle
\mathcal{L} = \mathcal{L}_{\mathrm{repair}} + \mathcal{L}_{\mathrm{label}} + \mathcal{L}_{\mathrm{recon}},
$}
\end{equation}
enabling the network to simultaneously restore disrupted structures, achieve accurate anatomical labeling, and reconstruct complete lung segments within a coherent optimization framework.

\section{Experiments and Results \label{sec:experiment}}
\begin{table*}[!t]
\centering
\caption{
\textbf{Quantitative comparison on the Lung3D+ dataset across airway, artery, and vein for pulmonary tree modeling.}
Voxel-based baselines operate on dense volumetric grids, while implicit function-based methods leverage structural priors for tree modeling. Overall, our TopoField consistently achieves the best performance across repair, labeling, and reconstruction tasks with efficient inference. Statistical significance tests on the repair task show that TopoField significantly outperforms the baselines ($p<0.05$). For IPGN, the reported total inference time is decomposed into labeling, reconstruction, and repair stages.
}
\label{tab:main-results}
\resizebox{\textwidth}{!}{
\renewcommand{\arraystretch}{1.2}
\begin{tabular}{lcccccccccc}
\toprule \toprule
\multirow{2}{*}{Method}
& \multicolumn{4}{c}{Repair} 
& \multicolumn{2}{c}{Labeling} 
& Reconstruction 
& \multirow{2}{*}{Inference Time (s)} \\
\cmidrule(lr){2-5} \cmidrule(lr){6-7} \cmidrule(lr){8-8}
& $CF1 \uparrow$ & $DMF1 \uparrow$ & $GDice \uparrow$ & $NCC \downarrow$ 
& $Dice_{Tree} \uparrow$ & $Dice_{Skeleton} \uparrow$ 
& $Dice_{Lung} \uparrow$ &  \\
\midrule
\rowcolor{lightgray}
\multicolumn{9}{c}{\textsc{Airway}} \\
\multicolumn{9}{l}{\textit{Voxel-based Baselines}} \\
3D-UNet (sliding window) 
& 77.98 & 67.32 & 71.48 & 4.50 
& 84.99 & 83.83 
& 48.11 & 67.35 \\
3D-UNet (down-sampled)   
& 12.95 & 15.91 & 41.92 & 8.01 
& 78.99 & 71.07 
& 53.32 & 0.01 \\
$G_{reco}$ (3D-UNet)  
& 78.18 & 68.65 & 71.72 & 4.25 
& -- & -- 
& -- & 108.00 \\
\hdashline
\multicolumn{9}{l}{\textit{Implicit Function-based Baselines}} \\
IPGN (Tree-6k) 
& -- & -- & -- & -- 
& 90.58 & 91.14 
& 80.54 & 11.59 {\footnotesize(0.22+11.37)} \\
IPGN (Surface-25k) 
& -- & -- & -- & -- 
& 90.32 & 90.79 
& 80.67 & 47.35 {\footnotesize(0.97+46.38)} \\
IPGN$^{\dagger}$ (Surface-25k)
& 0 & 0 & 0 & 15.10 
& 90.02 & 90.14 
& 80.74 & 183.66 {\footnotesize(0.97+46.38+136.31)} \\
\hdashline
TopoField (w/o Repair)
& -- & -- & -- & -- 
& \textbf{90.68} & \textbf{91.48} 
& \textbf{82.44} & 0.59 \\
TopoField 
& \textbf{85.48} & \textbf{84.09} & \textbf{79.33} & \textbf{3.44} 
& 89.16 & 90.18 
& 80.75 & 1.34 \\
TopoField (w/o GT)
& 77.85 & 76.61 & 76.20 & 4.87 
& 89.64 & 89.87 
& 77.65 & 1.34 \\


\midrule
\rowcolor{lightgray}
\multicolumn{9}{c}{\textsc{Artery}} \\
\multicolumn{9}{l}{\textit{Voxel-based Baselines}} \\
3D-UNet (sliding window) 
& 58.67 & 50.98 & 62.76 & 6.89 
& 85.69 & 80.37 & 55.12 
& 67.52 \\
3D-UNet (down-sampled)   
& 27.66 & 24.82 & 45.06 & 10.32 
& 74.21 & 70.08 & 62.15 
& 0.01 \\
$G_{reco}$ (3D-UNet)   
& 59.11 & 52.38 & 63.21 & 6.38 
& -- & -- 
& -- & 107.48 \\
\hdashline
\multicolumn{9}{l}{\textit{Implicit Function-based Baselines}} \\
IPGN (Tree-6k) 
& -- & -- & -- & -- 
& 89.79 & 90.39 & 81.65 
& 11.96 {\footnotesize(0.48+11.48)} \\
IPGN (Surface-25k) 
& -- & -- & -- & -- 
& 89.68 & 90.47 & 82.37 
& 48.57 {\footnotesize(2.09+46.48)} \\
IPGN$^{\dagger}$ (Surface-25k)
& 0 & 0 & 0 & 17.09 
& 85.55 & 85.22 & 79.33 
& 175.89 {\footnotesize(2.09+46.48+127.32)} \\
\hdashline
TopoField (w/o Repair)
& -- & -- & -- & -- 
& 89.45 & 89.87 
& 82.20 & 0.61 \\

TopoField 
& \textbf{68.31} & \textbf{66.78} & \textbf{71.61} & \textbf{5.06} 
& 88.88 & 88.99 & 81.07 
& 1.59 \\

TopoField (w/o GT) 
& 60.83 & 59.06 & 68.21 & 6.78 
& \textbf{90.19} & \textbf{91.05} & \textbf{82.67} 
& 1.59 \\


\midrule
\rowcolor{lightgray}
\multicolumn{9}{c}{\textsc{Vein}} \\
\multicolumn{9}{l}{\textit{Voxel-based Baselines}} \\
3D-UNet (sliding window) 
& 60.83 & 52.51 & 64.16 & 6.70 
& 79.96 & 70.33 & 58.13 
& 67.49 \\
3D-UNet (down-sampled)   
& 2.98 & 9.47 & 33.72 & 16.13 
& 77.72 & 67.61 & 62.18 
& 0.01 \\
$G_{reco}$ (3D-UNet)   
& 61.25 & 54.54 & 64.57 & 6.14 
& -- & -- 
& -- & 105.31 \\
\hdashline
\multicolumn{9}{l}{\textit{Implicit Function-based Baselines}} \\
IPGN (Tree-6k) 
& -- & -- & -- & -- 
& 83.72 & 78.43 & 77.01 
& 11.15 {\footnotesize(0.50+10.65)} \\
IPGN (Surface-25k) 
& -- & -- & -- & -- 
& 83.68 & 78.27 & 77.31 
& 48.70 {\footnotesize(2.22+46.48)} \\
IPGN$^{\dagger}$ (Surface-25k) 
& 0 & 0 & 0 & 17.48 
& 80.49 & 74.45 & 74.47 
& 175.83 {\footnotesize(2.22+46.48+127.13)}\\
\hdashline

TopoField (w/o Repair)
& -- & -- & -- & -- 
& \textbf{84.10} & \textbf{79.25} 
& \textbf{80.25} & 0.60 \\

TopoField 
& \textbf{66.93} & \textbf{66.05} & \textbf{71.88} & \textbf{5.90} 
& 83.35 & 78.50 & 79.21 
& 1.60 \\

TopoField (w/o GT) 
& 62.30 & 61.26 & 70.32 & 6.70 
& 83.52 & 78.63 & 79.52 
& 1.63 \\
\bottomrule \bottomrule
\end{tabular}
}
\vspace{-0.5em}
\end{table*}

\begin{table}[t]
\centering
\caption{
Quantitative comparison with point completion baselines on the Lung3D+ dataset.
The results show that general-purpose point completion models struggle to recover thin and topology-sensitive pulmonary tree structures.
}
\label{tab:point_completion_baselines}
\resizebox{\columnwidth}{!}{
\begin{tabular}{lcccc}
\toprule \toprule
Method 
& $CF1 \uparrow$ 
& $DMF1 \uparrow$ 
& $GDice \uparrow$ 
& $NCC \downarrow$ \\
\midrule

\rowcolor[gray]{0.90}
\multicolumn{5}{c}{Airway} \\
AdaPoinTr     & 1.00 & 0.07 & 7.09 & 3058 \\
AnchorFormer  & 0.51 & 0.03 & 1.41 & 3956 \\

\midrule
\rowcolor[gray]{0.90}
\multicolumn{5}{c}{Artery} \\
AdaPoinTr     & 0.21 & 0.05 & 0.49 & 6149 \\
AnchorFormer  & 0.29 & 0.06 & 0.85 & 4588 \\

\midrule
\rowcolor[gray]{0.90}
\multicolumn{5}{c}{Vein} \\
AdaPoinTr     & 0.07 & 0.01 & 0.31 & 6643 \\
AnchorFormer  & 0.13 & 0.02 & 0.48 & 5604 \\

\bottomrule \bottomrule
\end{tabular}
}
\end{table}

\begin{table*}[t]
\centering
\caption{
\textbf{Ablation study on topology repair.}
Quantitative comparison of IPGN$^{\dagger}$ and progressively enhanced TopoField variants on pulmonary tree repair across airway, artery, and vein datasets. For IPGN, the NCC reflects the degree of topological disconnection in the input tree. Results highlight the contribution of each proposed component to effective topology repair.
}
\label{tab:albation-repair}
\resizebox{\textwidth}{!}{
\begin{tabular}{lcccccccccccc}
\toprule \toprule
\multirow{2}{*}{Method}
& \multicolumn{4}{c}{Airway} 
& \multicolumn{4}{c}{Artery} 
& \multicolumn{4}{c}{Vein}  \\
\cmidrule(lr){2-5} \cmidrule(lr){6-9} \cmidrule(lr){10-13}
& $CF1 \uparrow$ & $DMF1 \uparrow$ & $GDice \uparrow$ & $NCC \downarrow$ 
& $CF1 \uparrow$ & $DMF1 \uparrow$ & $GDice \uparrow$ & $NCC \downarrow$ 
& $CF1 \uparrow$ & $DMF1 \uparrow$ & $GDice \uparrow$ & $NCC \downarrow$  \\
\midrule

IPGN$^{\dagger}$ 
& 0 & 0 & 0 & 15.10 
& 0 & 0 & 0 & 17.09
& 0 & 0 & 0 & 17.48 \\
\midrule
\rowcolor{lightgray}
\multicolumn{13}{c}{\textsc{Ours}} \\
TopoField (Base) 
& 38.64 & 37.71 & 59.63 & 14.34 
& 24.81 & 22.44 & 46.07 & 30.14
& 21.83 & 19.83 & 46.60 & 38.99 \\
TopoField (+SP) 
& 75.04 & 73.75 & 75.73 & 4.58 
& 61.90 & 60.93 & 69.19 & 5.60
& 57.63 & 55.82 & 69.53 & 7.89 \\
TopoField (+SP + Skel-GNN) 
& 75.97 & 73.98 & 76.36 & 4.71 
& 57.88 & 56.87 & 68.79 & 8.23
& 57.98 & 55.67 & 70.01 & 9.25 \\
TopoField (Full) 
& \textbf{85.48} & \textbf{84.09} & \textbf{79.33} & \textbf{3.44} 
& \textbf{68.31} & \textbf{66.78} & \textbf{71.61} & \textbf{5.06} 
& \textbf{66.93} & \textbf{66.05} & \textbf{71.88} & \textbf{5.90} \\
\bottomrule \bottomrule
\end{tabular}
}
\vspace{0.5em}
\end{table*} 

\subsection{Baselines}

We evaluate our proposed method on the Lung3D+ dataset against a variety of baseline approaches, using the evaluation metrics we designed to assess both reconstruction and labeling performance. The baselines cover three categories: voxel-based CNN methods, generic point completion methods, and implicit function-based methods. Unless otherwise specified, all experiments take as input a dense volume containing a disconnected pulmonary tree.

\noindent\textbf{Voxel-based CNN methods.}  
We adopt 3D CNNs to directly process dense volumetric inputs, where the task involves simultaneously reconstructing disconnected regions and performing semantic labeling of the pulmonary tree. Specifically, we employ a 3D U-Net with multiple output channels: one channel predicts the occupancy values of disconnected points for reconstruction, supervised by a combination of BCE and Dice loss, while the other channels perform semantic segmentation trained with a multi-class cross-entropy loss. We implement two variants of CNN-based methods.  

(1) \emph{Sliding-window based.} During training, we randomly select a known disconnection and crop a $64 \times 64 \times 64$ patch centered around it. The model is trained on such patches, and during inference, predictions are obtained using a sliding-window strategy with a stride of 32 across the full volume, and the results are aggregated.  

(2) \emph{Downsampling based.} The full volume is downsampled to $128^3$ voxels for training and the predictions are upsampled back to the original resolution during inference using nearest-neighbor interpolation. The downsampling not only reduces memory and computational cost but also preserves critical structural details. Notably, resolutions lower than $128^3$ were found to cause sparse disconnection points to vanish, leading to unstable training and convergence issues.  

We also considered other related vascular connectivity restoration methods, including MaskVSC~\citep{zhou2025masked} and CorSegRec~\citep{carneiro2024plug}. However, MaskVSC is designed for 2D retinal fundus images, whereas our task focuses on 3D pulmonary tree repair from binary masks. CorSegRec follows a multi-stage reconstruction pipeline with a different problem formulation, requiring intermediate steps such as disconnected-region identification and skeleton/lumen reconstruction. Since their implementations are not fully publicly available, faithful adaptation and fair comparison are difficult in our setting. Therefore, we discuss them as related methods rather than include them as direct baselines. In contrast, we adapt the learned post-processing model $G_{reco}$~\citep{carneiro2024restoring}, implemented with a 3D U-Net, as an additional CNN baseline that further refines the repaired output of the 3D U-Net baseline.

\noindent\textbf{Generic point completion methods.}
To further examine whether general-purpose point cloud completion models can address pulmonary tree repair, we additionally include two representative point completion baselines: AdaPoinTr~\citep{adapointr} and AnchorFormer~\citep{anchorformer}. These methods are originally designed to recover complete object shapes from partial point clouds, and have shown strong performance in generic 3D point completion benchmarks. For adaptation to our setting, we use the same input-output formulation as our method: the models take surface point clouds sampled from disconnected pulmonary trees as input and predict point sets corresponding to the missing regions. The predicted points are then evaluated under the same repair metrics as other baselines. However, unlike our topology-aware framework, these generic completion models do not explicitly model continuous spatial queries, semantic labeling, or anatomical tree structures. They are therefore included to examine the limitations of general point completion priors in repairing thin, sparse, and highly branched pulmonary trees.

\noindent\textbf{Implicit function-based methods.} 
The topology repair task requires the model to perform inference at arbitrary spatial locations, particularly in background regions where no explicit point samples are available. Conventional point-based networks are inherently limited in this regard, as they can only make predictions at observed point locations. In contrast, implicit function–based models enable continuous inference over the entire spatial domain, making them well suited for topology repair. Accordingly, we adopt IPGN~\citep{xie2025efficient}, a state-of-the-art implicit function–based framework for pulmonary tree labeling, as a representative baseline. While the original IPGN is designed for complete and well-connected tree structures, we adapt it to the more challenging setting of disconnected trees, where both point sets and graph representations are constructed from corrupted inputs.

Since IPGN is originally designed exclusively for the labeling task, we first evaluate two labeling-only variants. The first variant follows the original IPGN configuration by sampling 6k points from the tree structure, denoted as IPGN (Tree-6k). The second variant, denoted as IPGN (Surface-25k), samples 25k points from the tree surface to align the input representation with that used in our framework. Both variants adopt the same multi-stage training strategy as in the original IPGN implementation.

To further ensure a fair and comprehensive comparison, we additionally extend IPGN to handle the tree repair task by incorporating our proposed repair query sampling strategy. Specifically, repair query features are interpolated using the same mechanism employed for labeling queries in IPGN, and both the output format and loss functions are aligned with those of our framework. This extended variant also takes 25k surface points as input and is referred to as IPGN$^{\dagger}$ (Surface-25k). Despite these adaptations, IPGN$^{\dagger}$ remains limited in its ability to perform effective tree repair, as demonstrated in our experiments.

For comparison with the above baselines, we evaluate three variants of our proposed method. First, the full model, denoted as \textit{TopoField}, represents both surface and skeleton uniformly as point clouds and processes them using separate point-based encoders within a unified modeling framework. This design yields a compact yet expressive architecture that enables joint pulmonary tree repair, anatomical labeling, and lung segment reconstruction within a single forward pass.

In addition, we introduce a variant of the full model that disables the repair branches and performs tree and lung labeling exclusively, termed \textit{TopoField (w/o Repair)}. This variant is designed to closely align with the task scope of \textit{IPGN (Surface-25k)} for a controlled comparison. Finally, we evaluate \textit{TopoField (w/o GT)} under an incomplete topological supervision setting to assess the robustness and generalization capability of the proposed framework when dense ground-truth supervision is unavailable.

For the \textit{TopoField} with repair, we select the checkpoint according to the best repair performance, i.e., GDice, since topology repair is the primary objective of this work. Under this repair-oriented selection criterion, the corresponding labeling performance may fluctuate and is not necessarily optimal. In contrast, \textit{TopoField (w/o Repair)} performs only anatomical labeling, and its checkpoint is selected based on the best labeling performance. Therefore, it is expected that \textit{TopoField (w/o Repair)} can achieve slightly better labeling metrics. This difference mainly reflects the checkpoint-selection criterion rather than a substantial negative effect of the repair branch on labeling.

\subsection{Implementation Details}

In TopoField, the number of surface points is set to $N_s=25$k, which sufficiently preserves the global topology of the pulmonary tree and the locations of structural disconnections. The number of skeleton points is set to $N_k=6$k for airway data and increased to $N_k=10$k for the more complex artery and vein structures. Both surface and skeleton features are embedded in a $d=64$ dimensional space.

For the Super-Point Descriptor, the neighborhood size is fixed to $r=2$. In the Surface-to-Skeleton Attention module, each surface point attends to its ($K=8$) nearest skeleton neighbors, which empirically balances computational efficiency and local skeleton-context aggregation. In the Tri-Plane Projection Encoder, the resolution of each projected feature map is set to $R=256$, with a channel dimension of 64. The tri-plane representation stores learned skeleton-aware features rather than raw 2D projections, and query points sample from three planes with positional encoding to reduce bifurcation ambiguity. We set $R=256$ to balance distal-branch detail preservation with memory and computational cost. The positional encoding of query points also has an output dimension of 64.

For query sampling, the numbers of repair, labeling, and segmentation queries are all set to $Q_r=Q_l=Q_s=6$k. For repair queries, a sampling probability of $p=0.8$ is used, such that 80\% of queries are sampled near disconnections within a fixed radius of six times the corresponding branch radius. Under the weakly supervised setting, a \emph{single} additional disconnection is synthetically introduced on each corrupted tree using TopoBreak to provide implicit supervision.

The weights of the binary cross-entropy and Dice losses in $\mathcal{L}_{\mathrm{repair}}$ are both set to 0.5. TopoField is trained for 15 epochs with a batch size of 16 using the Adam optimizer and a learning rate of $1\times10^{-4}$. All models, including baselines, are implemented in PyTorch and trained on a single NVIDIA RTX 4090 GPU.

\subsection{Model Performance Comparison}

Table~\ref{tab:main-results} reports a comprehensive comparison across three pulmonary structures and three tasks: topology repair, labeling, and lung segment reconstruction. All results are averaged over the full test set. In addition to the metrics introduced in Section~3.4, we report the average per-case inference time to assess computational efficiency. Overall, our method consistently achieves superior repair performance, a more balanced accuracy across tasks, and a substantially better efficiency--accuracy trade-off than competing approaches.

\paragraph{Voxel-based CNN baselines}
Among voxel-based CNN methods, sliding-window inference consistently outperforms down-sampled variants in topology repair. Preserving dense local voxel information enables accurate detection of disconnections; however, this advantage does not fully translate to reconstruction quality. Notably, sliding-window methods exhibit a clear gap between CF1 and DMF1, indicating that although disconnections are often detected, the repaired regions poorly match the true tubular geometry. In contrast, down-sampled approaches, while computationally efficient, severely distort fine-scale structures, weaken reconstruction supervision, and degrade repair accuracy. Interestingly, they perform relatively well in lung segment reconstruction, suggesting that this task relies more on global lung modeling than on precise local topology.

\begin{figure}[tb]
    \centering
	\includegraphics[width=0.9\linewidth]{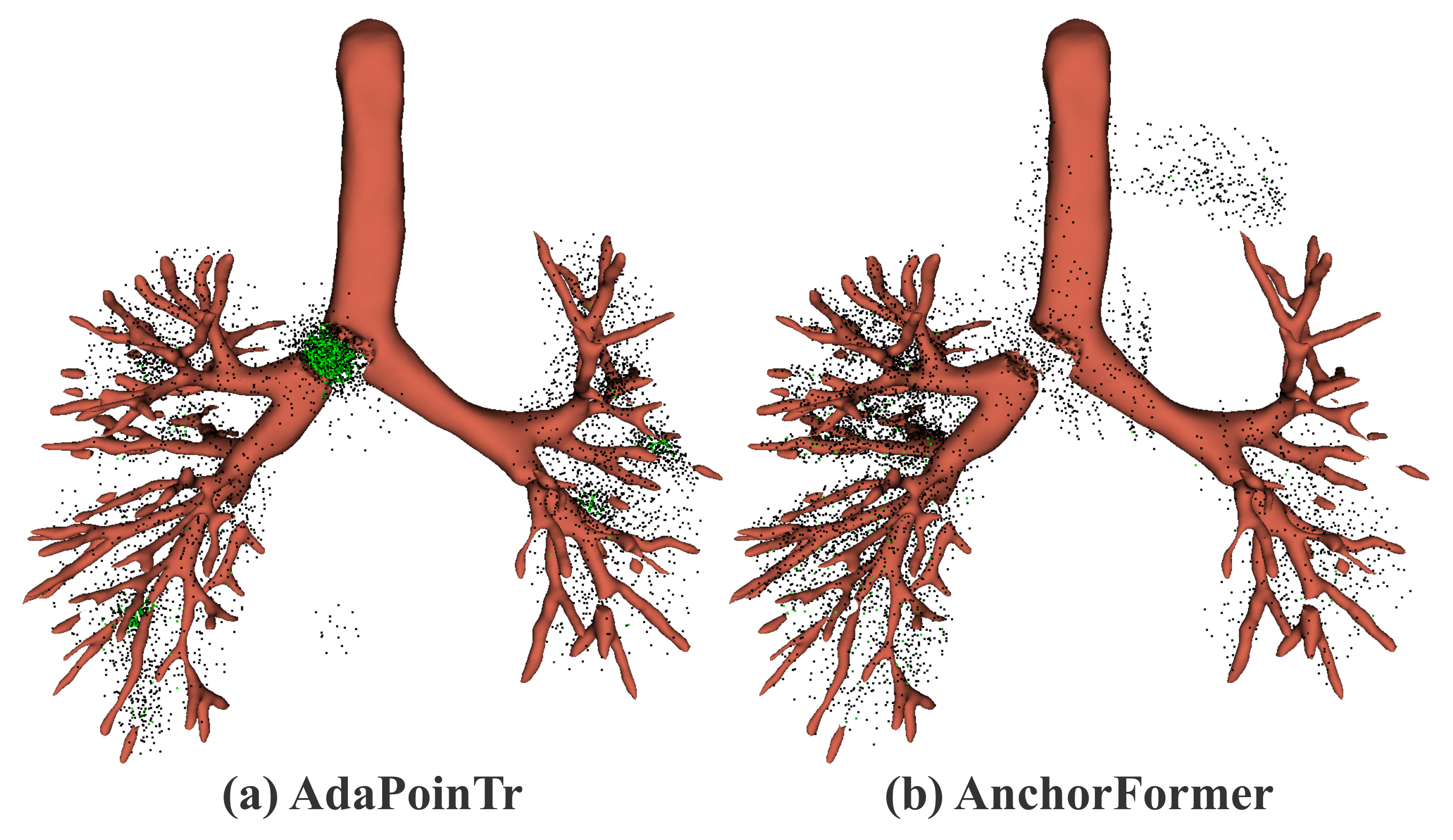}
	\caption{\textbf{Failure cases of generic point completion baselines on pulmonary tree repair.} 
    AdaPoinTr and AnchorFormer predict spatially scattered point sets and fail to recover thin distal branches, leading to poor voxel-level repair and limited connectivity restoration.
    }
	\label{fig:point_completion_vis}
\end{figure}

\paragraph{Point completion baselines}
As shown in Table~\ref{tab:point_completion_baselines}, both AdaPoinTr and AnchorFormer perform poorly on the pulmonary tree repair task. Across airway, artery, and vein, their voxel-level repair scores remain consistently low, with CF1, DMF1, and GDice all below 7.09\%, while the NCC values are substantially higher than those of topology-aware repair methods. The qualitative results in Fig.~\ref{fig:point_completion_vis} show that the predicted points are scattered around the tree surface and rarely form continuous repaired branches, particularly in thin distal regions. These results indicate that, despite using the same input-output setting as our method, generic point completion models are not well suited to fine-grained pulmonary tree repair, where accurate localization of multiple small disconnected components is more critical than generating a globally plausible point set.

\paragraph{IPGN family}
IPGN variants achieve strong labeling performance but fail to converge on the repair task. IPGN$^\dagger$ produces low repair probabilities with no positive voxels after thresholding, indicating its difficulty in recovering small, dispersed missing tubular regions from disconnected inputs. This limitation is likely caused by nearest-neighbor feature interpolation, which struggles to distinguish query points in locally disconnected regions, where interpolated features from occupied and unoccupied voxels become ambiguous. Such ambiguity is less detrimental for labeling, as neighboring branches typically share the same anatomical class. Moreover, IPGN depends on a complete skeleton graph; skeleton disconnections disrupt graph connectivity and impair feature learning at critical local sites. In addition, inference time increases sharply with the number of input points due to nearest-neighbor queries, limiting scalability. Notably, IPGN and IPGN (Surface-25k) do not include a repair task and therefore report inference time only for labeling and lung segment reconstruction. 

\paragraph{Overall performance and metric sensitivity of TopoField}
Our full model achieves state-of-the-art performance on topology repair while maintaining competitive labeling and lung segment reconstruction accuracy compared with IPGN. Crucially, TopoField exhibits closely matched CF1 and DMF1 scores, indicating not only reliable detection of disconnections but also accurate reconstruction of anatomically plausible tubular structures. This behavior contrasts sharply with voxel-based CNNs and demonstrates TopoField’s ability to learn meaningful geometric priors of disconnected pulmonary trees. In addition, the labeling-only variant, \textit{TopoField (w/o Repair)}, achieves labeling performance comparable to IPGN (Surface-25k), with slight labeling improvements on airway and vein samples, while consistently demonstrating superior segment reconstruction across all three structures. This result underscores the effectiveness of the unified implicit representation even when applied to a single task. Meanwhile, TopoField enables real-time inference, with an average processing time slightly above one second per case, substantially outperforming both IPGN and voxel-based baselines in efficiency. Finally, \textit{TopoField (w/o GT)} maintains strong performance across all tasks with only a modest reduction in repair accuracy, demonstrating practical robustness in realistic scenarios where explicit disconnection supervision is unavailable.

\paragraph{Structural complexity and generalization}
Performance on artery and vein datasets is consistently lower than on airways, reflecting their higher geometric complexity and more intricate branching patterns. Nevertheless, the relative performance trends remain consistent across airway, artery, and vein datasets, underscoring the robustness and generalizability of our framework across different pulmonary structures.

\begin{figure*}[tb]
    \centering
	\includegraphics[width=0.95\linewidth,height=0.34\textheight]{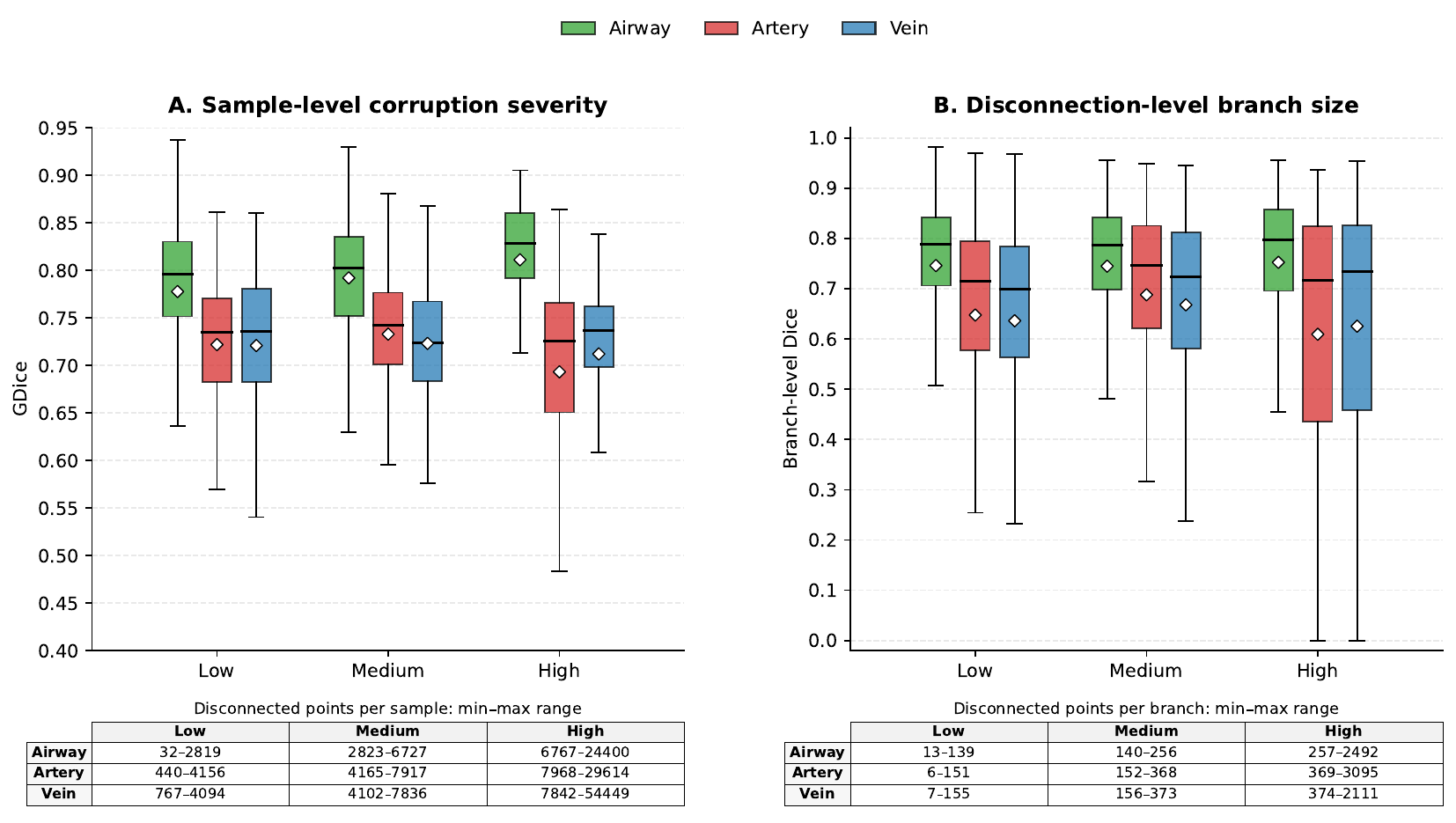}
	\caption{\textbf{Corruption-severity sensitivity analysis of repair performance.} 
    Repair performance is evaluated under different corruption severity levels for airway, artery, and vein. (A) Sample-level GDice across low, medium, and high corruption severity groups defined by the number of disconnected points per sample. (B) Branch-level Dice across low, medium, and high branch-size groups defined by the number of disconnected points per branch. Low, medium, and high groups are determined separately for each structure using tertiles. The tables below each panel report the corresponding min–max range of disconnected points for each group.
    }
	\label{fig:corruption_severity}
\end{figure*}

\subsection{Corruption-Severity Sensitivity Analysis}

To evaluate the robustness of TopoField under different disconnection severities, we conduct a corruption-severity sensitivity analysis on the airway, artery, and vein test sets.

We perform the analysis at two levels. At the sample level, corruption severity is measured by the number of disconnected points in each sample, and repair performance is evaluated using GDice. At the disconnection level, branch size is measured by the number of disconnected points in each disconnected branch, and performance is evaluated using the matched branch-level Dice. For each structure, samples and disconnected branches are divided into low-, medium-, and high-severity groups using tertiles. For the disconnection-level analysis, branches larger than the structure-specific 95th percentile are excluded before grouping to reduce the influence of extreme outliers.

As shown in Fig.~\ref{fig:corruption_severity}, TopoField maintains stable GDice across different sample-level severity groups for all three structures, indicating that the global repair quality is robust to increasing corruption severity. At the disconnection level, TopoField achieves consistently good Dice scores for small branches, demonstrating its ability to repair fine distal disconnections. Although larger disconnected branches show higher variability, especially for artery and vein, the overall branch-level performance remains stable across severity groups. These results indicate that TopoField is robust to varying disconnection difficulties at both the sample and branch levels.

\subsection{Ablation Study on Topology Repair}

Motivated by the observation that IPGN fails to perform topology repair, this section investigates the key factors that enable effective repair in our framework and presents a systematic ablation study of the proposed components.

We evaluate four progressively enhanced variants of TopoField. The first variant, \textit{TopoField (Base)}, employs only the surface branch while retaining the unified implicit field, without incorporating super-point descriptors or skeleton information. The second variant, \textit{TopoField (+SP)}, augments the surface branch with the proposed super-point descriptor to enrich local structural context. The third variant, \textit{TopoField (+SP + Skel-GNN)}, further introduces an explicit skeleton branch represented as a graph and processed using a Graph Attention Network (GAT), following the design philosophy of IPGN, before fusion via the SSA module. Finally, \textit{TopoField (Full)} represents both surface and skeleton uniformly as point clouds, constituting the complete version of our framework.

As shown in Table~\ref{tab:albation-repair}, IPGN$^{\dagger}$ (Surface-25k) fails entirely on the topology repair task across all three anatomical structures, yielding zero CF1, DMF1, and GDice scores. In contrast, even the simplest variant, \textit{TopoField (Base)}, achieves non-trivial repair performance, demonstrating that embedding pulmonary tree features into a unified encoded implicit field already enables meaningful topology restoration. This highlights the importance of continuous, query-based inference, which allows fine-grained feature interpolation at arbitrary spatial locations and is unattainable with nearest-neighbor interpolation alone.

Introducing super-point descriptors in \textit{TopoField (+SP)} leads to a substantial performance boost across all structures, with CF1 improving from 38.64 to 75.04 on airways and from 24.81 to 61.90 on arteries, alongside a dramatic reduction in NCC. These gains indicate that enriching local structural context is critical for accurately recovering disconnected branches, particularly in regions with complex geometry. Interestingly, incorporating an explicit skeleton graph via \textit{TopoField (+SP + Skel-GNN)} does not yield consistent improvements and even degrades performance on arteries and veins, as reflected by lower CF1 and higher NCC values compared with \textit{TopoField (+SP)}. This suggests that graph-based representations are sensitive to connectivity disruptions and may propagate erroneous information when the underlying topology is corrupted. By contrast, \textit{TopoField (Full)} consistently achieves the best repair performance across all metrics and anatomical structures, attaining CF1 scores of 85.48, 68.31, and 66.93 on airway, artery, and vein data, respectively. Representing skeletons as point clouds avoids reliance on explicit graph connectivity and enables effective surface-to-skeleton feature fusion through local cross-attention, resulting in more accurate and anatomically plausible reconstructions.

Overall, the superior topology repair capability of TopoField can be attributed to three key factors: (1) a unified encoded implicit field that supports continuous and fine-grained query-based inference, (2) a unified point-based representation of surface and skeleton features fused via local cross-attention, and (3) enriched local structural modeling through super-point descriptors.

\begin{figure*}[tb]
    \centering
	\includegraphics[width=\linewidth]{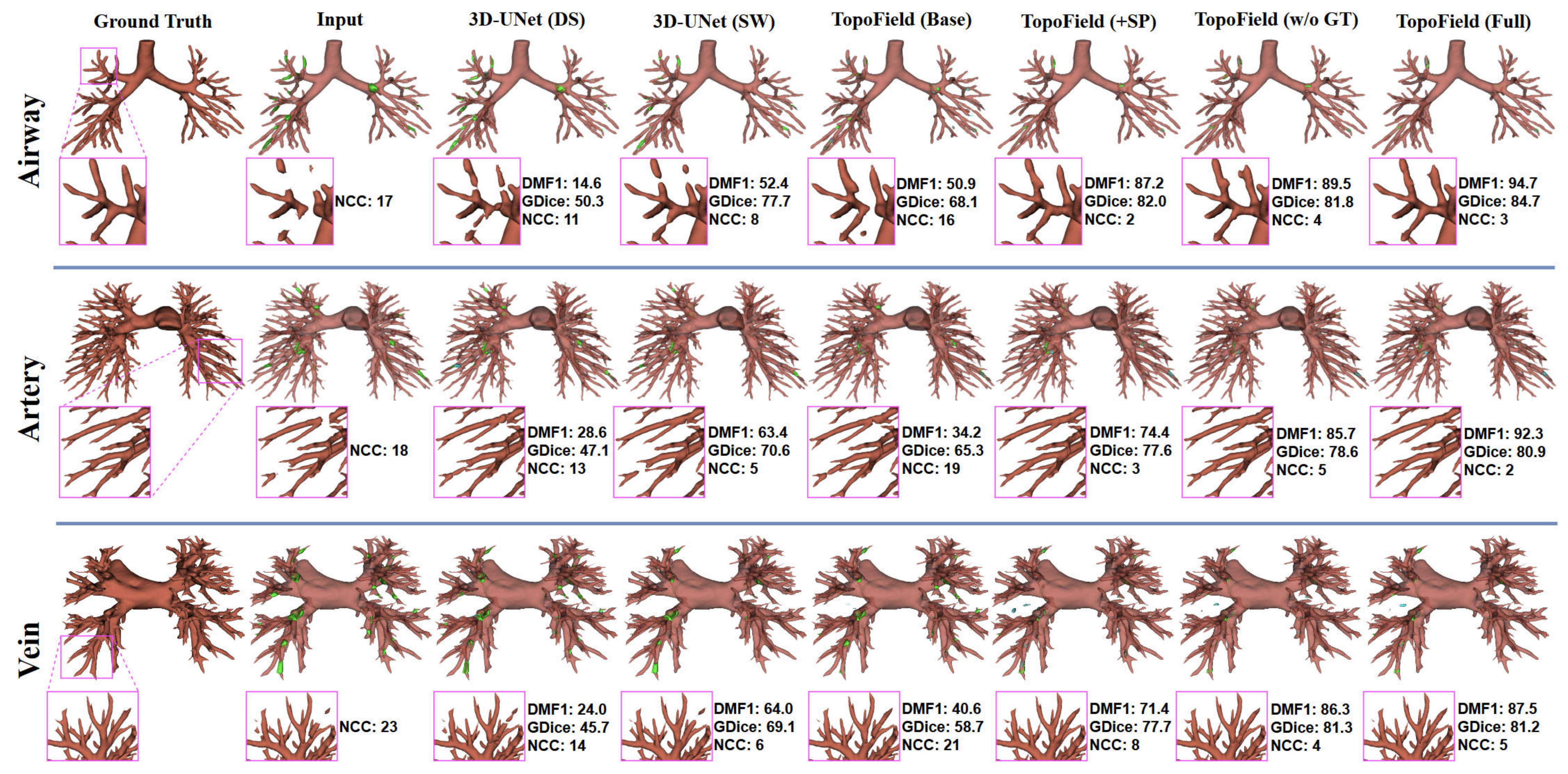}
	\caption{\textbf{Visualization of Pulmonary Tree Topology Repair Results.} 
    Representative samples from the airway, artery, and vein datasets (top to bottom) are shown for comparison. From left to right, each row presents the ground truth, corrupted input, two 3D-UNet baselines (DS: down-sampled; SW: sliding-window), and different TopoField variants. Target structures are rendered in red, false positives in blue, and false negatives (disconnected segments) in green. Purple boxes indicate locally disconnected regions, with zoomed-in views highlighting repair details. DMF1 (\%), GDice (\%), and NCC are reported for each method on the corresponding sample. Overall, TopoField variants, particularly TopoField (Full), achieve more accurate reconnections and better structural consistency with the ground truth than the 3D-UNet baselines across all three pulmonary structures.
    }
	\label{fig:fig-vis-1}
\end{figure*}

\begin{figure*}[tb]
    \centering
	\includegraphics[width=\linewidth]{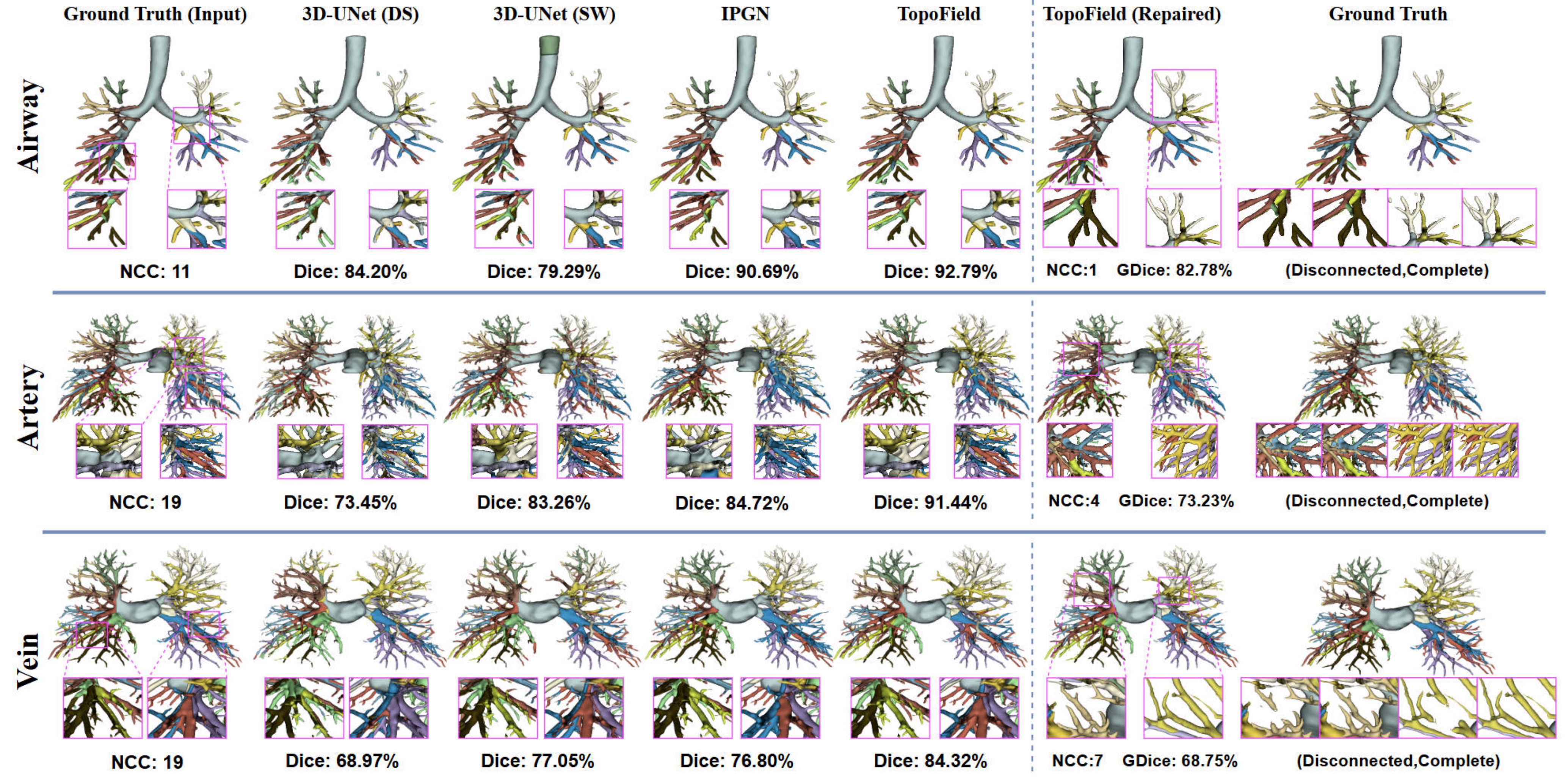}
	\caption{\textbf{Visualization of Pulmonary Tree Labeling Results.} 
    Qualitative comparison of 19-class labeling on airway, artery, and vein structures. From left to right, results are shown for the ground truth (input), two 3D-UNet baselines (DS: down-sampled; SW: sliding-window), IPGN (IPGN$^{\dagger}$), and TopoField. The first six columns present labeling on the disconnected input tree with input NCC and Dice scores, while the last two columns show labeling after TopoField-based topology repair and the complete ground truth, reporting NCC and GDice. For clarity, both disconnected and complete local patches are illustrated in the ground truth column. Purple boxes highlight local regions of interest with zoomed-in views.}
	\label{fig:fig-vis-2}
\end{figure*}

\begin{figure*}[tb]
    \centering
	\includegraphics[width=\linewidth]{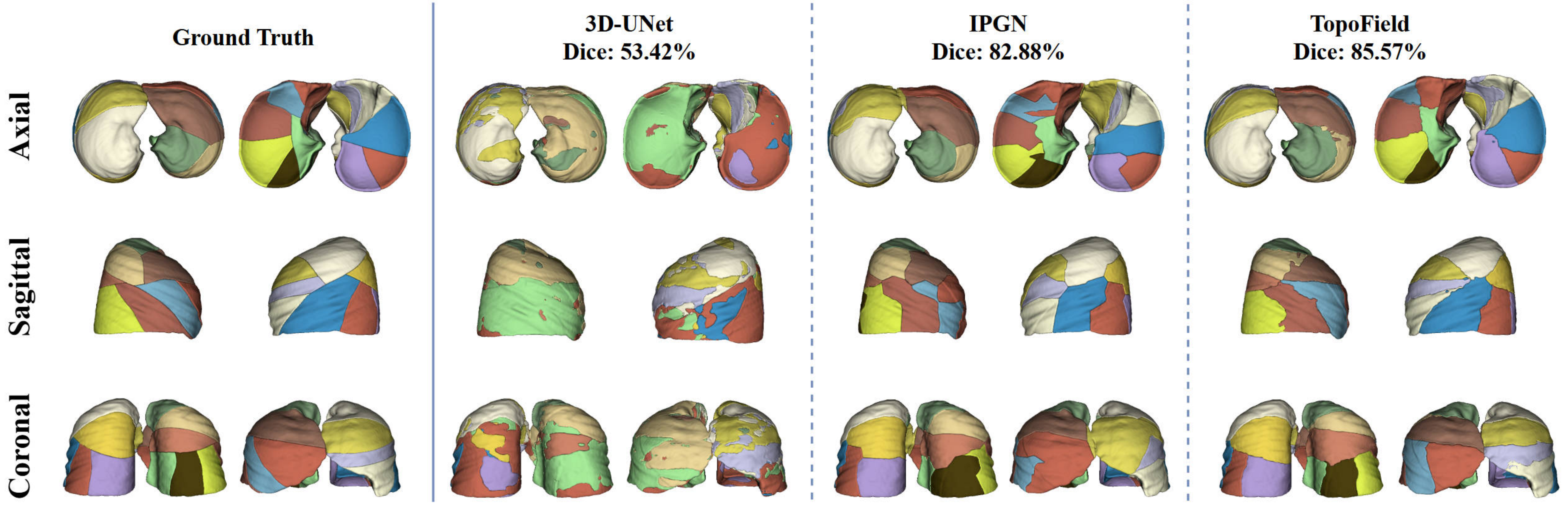}
	\caption{\textbf{Visualization of Lung Segment Reconstruction Results.}
    Qualitative comparison of 18-class lung segment reconstruction using 3D-UNet (down-sampled), IPGN, and TopoField, with all models trained on the airway dataset. For each method, reconstruction results from the axial, sagittal, and coronal views of the same sample are shown. The corresponding Dice score is reported for quantitative comparison.
    }
	\label{fig:fig-vis-3}
\end{figure*}

\subsection{Ablation Study on Surface-to-Skeleton Fusion}
\begin{table}[]
\centering
\renewcommand{\arraystretch}{1.15}
\caption{
\textbf{Ablation study on surface-to-skeleton feature fusion strategies on the \emph{airway} and \emph{artery} datasets.}
}
\label{tab:fusion-ablation}
\resizebox{\columnwidth}{!}{
\begin{tabular}{lcccccc}
\toprule \toprule
\multirow{2}{*}{Method}
& \multicolumn{3}{c}{Repair}
& \multicolumn{2}{c}{Labeling}
& Reconstruction \\
\cmidrule(lr){2-4} \cmidrule(lr){5-6} \cmidrule(lr){7-7}
& $DMF1 \uparrow$ & $GDice \uparrow$ & $NCC \downarrow$
& $Dice_{Tree} \uparrow$ & $Dice_{Skeleton} \uparrow$
& $Dice_{Lung} \uparrow$ \\
\midrule
\rowcolor{lightgray}
\multicolumn{7}{c}{\emph{Airway dataset}} \\
\midrule
TriPlane Fusion (Early)      & 72.61 & 75.88 & 4.87 & 86.97 & 85.30 & 77.03 \\
TriPlane Fusion (Late)       & 75.22 & 75.90 & 4.66 & 88.62 & 87.97 & 79.06 \\
Distance-weighted Fusion     & 70.33 & 74.88 & 5.07 & 87.73 & 86.52 & 77.23 \\
\rowcolor{lightgray}
\textbf{SSA (Ours)}          & \textbf{84.09} & \textbf{79.33} & \textbf{3.44} & \textbf{89.16} & \textbf{90.18} & \textbf{80.75} \\
\midrule
\rowcolor{lightgray}
\multicolumn{7}{c}{\emph{Artery dataset}} \\
\midrule
TriPlane Fusion (Early)      & 54.15 & 67.59 & 9.08 & 83.82 & 81.08 & 74.63 \\
TriPlane Fusion (Late)       & 58.17 & 67.25 & 7.89 & 85.07 & 83.67 & 77.40 \\
Distance-weighted Fusion     & 55.64 & 67.52 & 8.37 & 84.29 & 82.90 & 76.19 \\
\rowcolor{lightgray}
\textbf{SSA (Ours)}          & \textbf{66.78} & \textbf{71.61} & \textbf{5.06} & \textbf{88.88} & \textbf{88.99} & \textbf{81.07} \\
\bottomrule \bottomrule
\end{tabular}
}
\end{table}

This section investigates alternative surface–skeleton fusion strategies to validate the design of the proposed surface-to-skeleton attention (SSA) module. Beyond the local cross-attention mechanism in SSA, we implement three representative fusion variants while keeping the remaining pipeline unchanged.

\paragraph{(1) Early point-level fusion before projection}
In the first variant (TriPlane Fusion (Early) in Table~\ref{tab:fusion-ablation}), surface and skeleton points are merged immediately after point encoding. Their features are jointly projected into tri-plane representations and processed by the U-Net, formulated as
\begin{equation}
\mathcal{F}
=
\text{UNet}\!\left(
\text{Proj}\!\left(\Phi_s \cup \Phi_k \mid \{\mathcal{S}, \mathcal{K}\}\right)
\right).
\end{equation}

\paragraph{(2) Late fusion at the tri-plane feature level}
In the second variant (TriPlane Fusion (Late) in Table~\ref{tab:fusion-ablation}), surface and skeleton features are projected independently into tri-plane representations and concatenated along the channel dimension before U-Net processing:
\begin{equation}
\mathcal{F}
=
\text{UNet}\!\left(
\text{Proj}(\Phi_s \mid \mathcal{S})
\oplus
\text{Proj}(\Phi_k \mid \mathcal{K})
\right).
\end{equation}

\paragraph{(3) Local distance-weighted point–graph fusion}
The third variant (Distance-weighted Fusion in Table~\ref{tab:fusion-ablation}) adopts the point–graph fusion strategy used in IPGN, replacing SSA’s local cross-attention with a distance-weighted aggregation of the $K$ nearest skeleton features for each surface point. The aggregated skeleton feature is concatenated with the corresponding surface feature and mapped back to the original feature dimension via a linear projection, while all subsequent stages of the pipeline remain unchanged. The key distinction from SSA therefore lies solely in the mechanism by which surface features aggregate information from the skeleton.

Table~\ref{tab:fusion-ablation} reports the ablation results of different surface-to-skeleton fusion strategies within TopoField on the airway and artery datasets. Early and late feature concatenation yield only marginal improvements, suggesting that naive fusion cannot effectively model the geometric correspondence between surface and skeleton, while distance-weighted fusion is limited by fixed distance-based weights. In contrast, SSA consistently achieves the best performance across repair, labeling, and reconstruction, demonstrating the effectiveness of adaptive, geometry-aware feature aggregation. The consistent trend on arteries further indicates that SSA generalizes to more complex pulmonary vascular structures.

\subsection{Visualization}

In this section, we present a qualitative comparison between our proposed method and representative baseline approaches. Using selected samples from the airway, artery, and vein datasets, we visually evaluate the performance of different methods across three core tasks: topology repair, tree labeling, and lung segment reconstruction.

Figure~\ref{fig:fig-vis-1} illustrates the topology repair performance of different TopoField variants and two 3D-UNet baselines on disconnected pulmonary trees. The IPGN method is excluded, as it fails to perform the repair task. To facilitate visual comparison, false positives and false negatives are highlighted, and zoomed-in views are provided for local regions of interest. While both TopoField and 3D-UNet variants can detect disconnections in distal branches, TopoField (+SP), TopoField (w/o GT), and TopoField (Full) achieve substantially improved repair quality, characterized by enhanced structural connectivity and lower NCC values. In particular, TopoField (Full) produces reconstructions that most closely match the ground truth morphology, as further supported by higher DMF1 and GDice scores.

Comparing TopoField (+SP) with TopoField (Base), the inclusion of the super-point descriptor leads to a clear performance gain, indicating its effectiveness in capturing richer local structural information for fine-grained reconstruction. In addition, artery and vein structures present greater repair challenges due to their increased structural complexity, including thinner branches and more diverse branching directions, resulting in lower overall metrics and more false positives. Despite these challenges, TopoField consistently outperforms the 3D-UNet baselines, successfully repairing the majority of disconnected regions.

Figure~\ref{fig:fig-vis-2} visualizes the 19-class pulmonary tree labeling performance of the two 3D-UNet baselines, IPGN, and TopoField. Overall, TopoField achieves superior labeling accuracy compared with the baseline methods. On the airway sample, TopoField performs comparably to IPGN, while on the artery and vein samples it exhibits noticeably improved labeling quality, consistent with the quantitative results reported in Table~\ref{tab:main-results}. 

Tree labeling is inherently challenging near intersegmental boundaries, where vessels or airways may be adjacent to multiple anatomical segments, leading to ambiguous class assignments. Both IPGN and TopoField yield consistent predictions within individual segments, and most misclassifications occur near these boundaries. Notably, TopoField demonstrates improved smoothness and inter-class consistency in boundary regions, resulting in more coherent multi-class labeling. Moreover, TopoField preserves accurate and consistent labels after topology repair, highlighting its ability to jointly address topology repair and tree labeling.

Finally, Figure~\ref{fig:fig-vis-3} presents the 18-class lung segment reconstruction results of 3D-UNet (down-sampled), IPGN, and TopoField from three anatomical views. For clarity, only the down-sampled 3D-UNet is included, as it outperforms its sliding-window counterpart. As shown, 3D-UNet struggles to maintain global continuity and inter-segment consistency, resulting in fragmented reconstructions, while IPGN still exhibits inaccuracies near segment boundaries. In contrast, TopoField accurately reconstructs lung segment boundaries and achieves the closest morphological agreement with the ground truth, consistent with its superior Dice score.

\begin{figure}[tb]
    \centering
	\includegraphics[width=\linewidth]{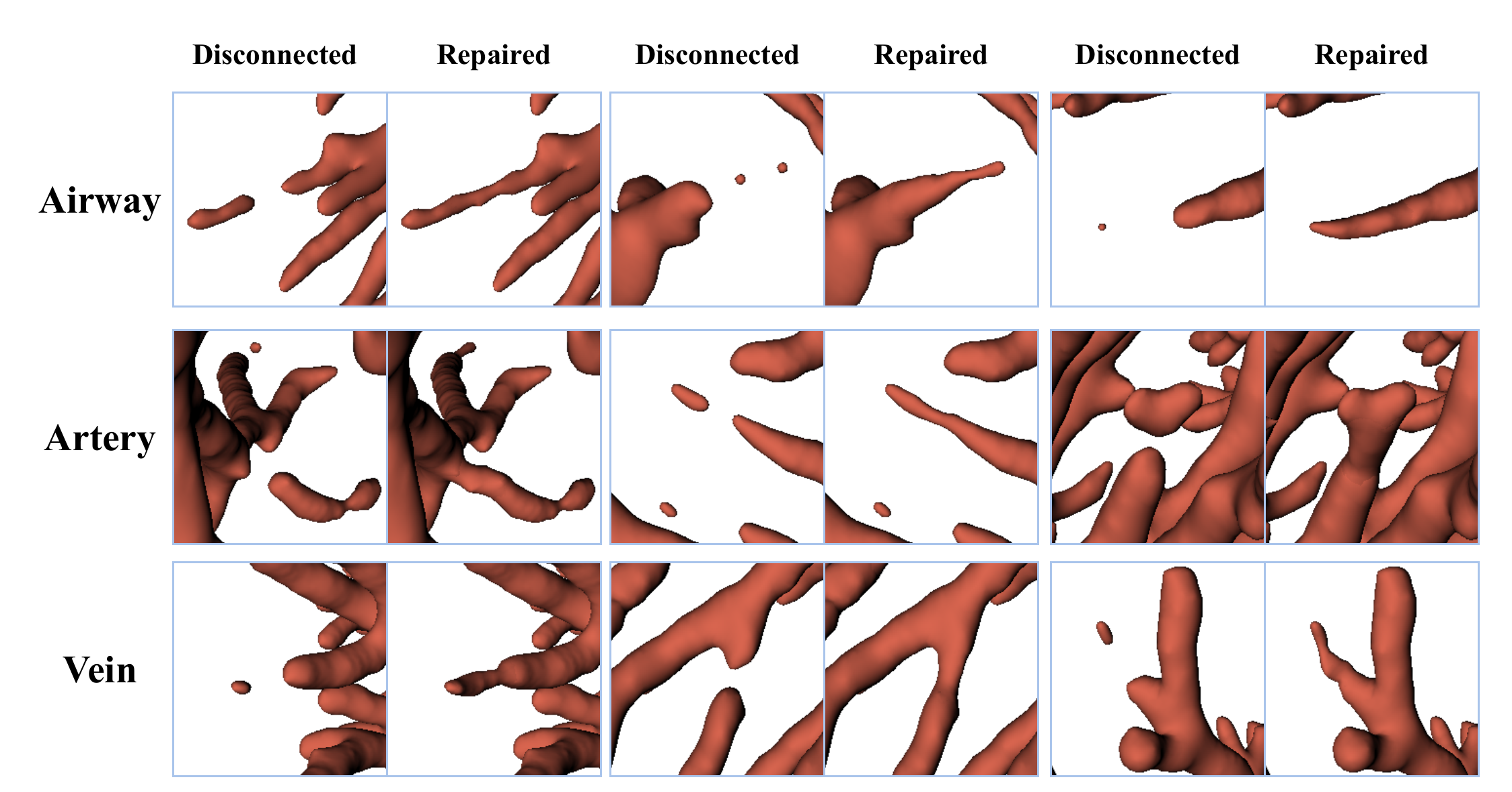}
	\caption{\textbf{Generalization to Real Segmentation Outputs.} TopoField repairs naturally occurring disconnections in TotalSegmentator segmentation results, demonstrating its applicability beyond synthetic corruptions.}
	\label{fig:real_segmentation_repair}
\end{figure}

\subsection{Qualitative Evaluation on Real Segmentation Disconnections}
\label{sec:real_segmentation}

To further assess the practical applicability of TopoField, we conduct an additional qualitative experiment on real segmentation outputs. Unlike TopoBreak-generated corruptions, real segmentation masks may contain naturally occurring disconnections and other segmentation errors.

Specifically, we apply TotalSegmentator to the CT scans from the Lung3D+ test set to obtain airway, artery, and vein segmentation masks, which are then used as inputs to TopoField. Since TotalSegmentator was not trained on Lung3D+, this setting serves as an external qualitative generalization test. Because the missing regions in these real segmentation outputs do not have voxel-wise annotations, quantitative reconstruction metrics cannot be reliably computed. Therefore, this experiment focuses on whether TopoField can reconnect local disconnected structures in imperfect segmentation masks, rather than claiming an overall improvement in segmentation accuracy.

As shown in Fig.~\ref{fig:real_segmentation_repair}, several small branches or fragments in the airway, artery, and vein masks are disconnected from the main anatomical tree. After applying TopoField, these structures are reconnected to nearby plausible branches, improving local topological continuity. This suggests that TopoField is not limited to synthetic TopoBreak corruptions, but can also repair real disconnections produced by an external segmentation model.

This experiment also reveals a limitation: TopoField is designed to recover missing connections, not to remove false-positive components already present in the input segmentation. Thus, TopoField should be viewed as a topology repair module complementary to segmentation models.

\begin{figure}[tb]
    \centering
	\includegraphics[width=\linewidth]{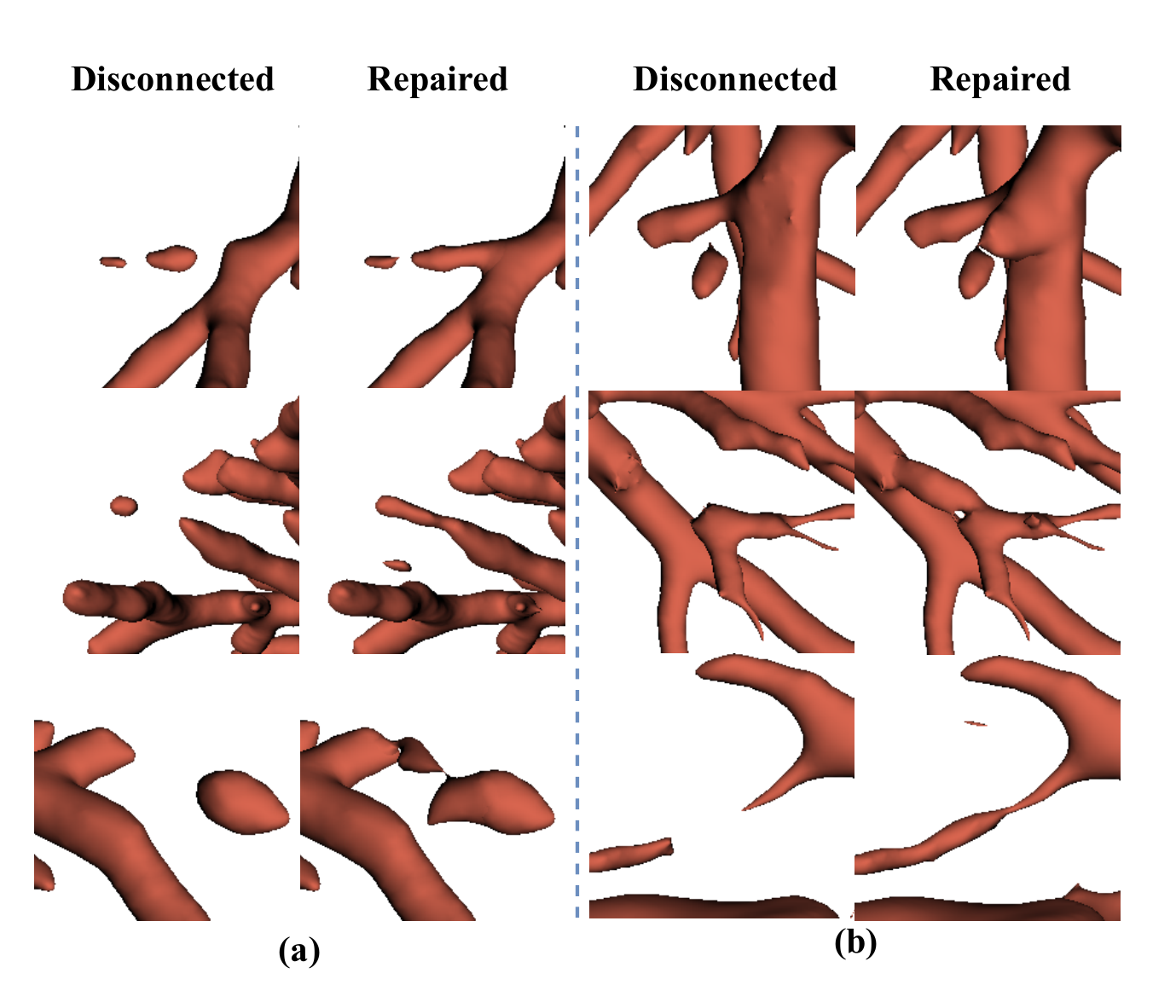}
	\caption{\textbf{Failure cases of TopoField.} (a) Real TotalSegmentator outputs showing incomplete reconnections, small spurious branches, or connections to isolated false-positive components. (b) Lung3D+ test cases showing incomplete reconnections or small spurious branches after repair.}
	\label{fig:failure_cases}
\end{figure}

\subsection{Failure Case Analysis}
\label{sec:failure_cases}

Although TopoField effectively restores most disconnected pulmonary tree structures, challenging cases remain. As shown in Fig.~\ref{fig:failure_cases}, two representative failure modes are observed. First, the model may produce incomplete reconnections, where only part of the missing structure is recovered. This typically occurs for thin distal branches, large gaps, or locally ambiguous geometries. Second, the model may generate small spurious branches near the repaired regions, resulting in false-positive local connections.

This limitation is more pronounced on real segmentation outputs, where the input masks may already contain isolated false-positive components. Since TopoField is trained to recover missing tubular structures, such components can be interpreted as disconnected branches and connected to nearby trees. This is mainly due to the current corruption simulation, which focuses on missing structures but does not explicitly model false-positive noise. These observations indicate that TopoField is better viewed as a topology repair module rather than a complete segmentation error correction framework.

\section{Discussion and Conclusion}
In this work, we present TopoField, a topology-aware implicit modeling framework that treats pulmonary tree repair as a first-class problem. The proposed method jointly models pulmonary tree geometry and topology via complementary surface–skeleton point cloud representations, enhanced by a super-point descriptor, a surface-to-skeleton attention mechanism, and a unified implicit field, enabling fine-grained and scalable topology restoration.

To improve robustness and practical applicability, we further propose a topological repair strategy without disconnection annotations, allowing the model to learn connectivity restoration under unknown and unlocalized breakpoints. By synthesizing additional disruptions on corrupted trees and learning to recover them, the proposed strategy eliminates the need for explicit topological supervision while remaining faithful to real-world deployment settings.

Building upon this repair-centric formulation, TopoField is extended to a unified multi-task implicit inference framework that simultaneously performs pulmonary tree repair, anatomical labeling, and lung segment reconstruction within a single forward pass. This unified design enables end-to-end optimization across tasks and achieves highly efficient inference, completing all three tasks in just over one second per case. Extensive experiments demonstrate that TopoField achieves accurate topology repair while delivering competitive semantic labeling and segment reconstruction performance. Additional comparisons with generic point completion methods and qualitative validation on real segmentation outputs further highlight its effectiveness and practical potential for clinical pulmonary tree analysis.

Despite the effectiveness of TopoField, several aspects remain to be further explored.

First, the main experiments in this work are primarily based on TopoBreak-simulated disconnections, which enable controlled and systematic evaluation of connectivity restoration. However, the current simulation mainly models missing or disconnected branches and does not explicitly simulate leakage artifacts, false-positive components, or spurious branches. Although our qualitative validation on real segmentation outputs shows that TopoField can reconnect naturally occurring disconnections, it also reveals challenging cases involving leakage, false positives, and small spurious branches. Moreover, this validation remains primarily qualitative, since real-error quantitative benchmarks with voxel-wise annotations of missing and spurious regions are currently unavailable. Future work will therefore extend both the training and evaluation settings to cover more realistic topological degradations, and explore tighter integration with upstream segmentation models to jointly handle missing connectivity and erroneous structures.

Second, for anatomical labeling and lung segment reconstruction, query points are currently sampled uniformly from the pulmonary tree and lung regions, without explicitly prioritizing anatomically critical boundary areas. In practice, classification ambiguity is most pronounced near intersegmental interfaces, where subtle geometric cues determine class transitions. Future work will therefore investigate boundary-aware query sampling strategies and complementary loss formulations that emphasize these regions, with the aim of improving sensitivity at anatomical boundaries and promoting smoother, more anatomically consistent labeling and reconstruction results.

\section{Acknowledgments}

Z.W. was supported by Australian Government Research Training Program (RTP) scholarship. J.Y. was supported by the ELLIS Institute Finland and School of Electrical Engineering, Aalto University. 

\bibliographystyle{model2-names.bst}\biboptions{authoryear}
\bibliography{main}

\end{document}